\documentclass{article}

\usepackage{arxiv}

\usepackage[utf8]{inputenc} 
\usepackage[T1]{fontenc}    
\usepackage{hyperref}       
\usepackage{url}            
\usepackage{booktabs}       
\usepackage{amsfonts}       
\usepackage{amsmath}
\usepackage{nicefrac}       
\usepackage{microtype}      
\usepackage{lipsum}
\usepackage{graphicx}
\usepackage{multicol}
\usepackage{multirow}
\usepackage{array}
\graphicspath{ {./images/} }

\def\ie{{\it i.e.}}
\def\eg{{\it e.g.}}
\def\etal{{\it et al.}}
\DeclareMathOperator*{\argmax}{argmax}

\newsavebox\CBox
\def\textbff#1{\sbox\CBox{#1}\resizebox{\wd\CBox}{\ht\CBox}{\textbf{#1}}}

\usepackage[caption=false,font=footnotesize]{subfig}

\usepackage{algorithm}
\usepackage[noend]{algpseudocode}

\title{Co-Training with Active Contrastive Learning and Meta-Pseudo-Labeling on 2D Projections for Deep Semi-Supervised Learning}

\author{
 David Aparco-Cardenas \\
  Institute of Computing\\
  University of Campinas\\
  Campinas, SP 13083-852, Brazil \\
  \texttt{david.cardenas@ic.unicamp.br} \\
  \And
 Jancarlo F. Gomes \\
  Institute of Computing\\
  University of Campinas\\
  Campinas, SP 13083-852, Brazil \\
  \texttt{jgomes@ic.unicamp.br} \\
  \And
 Alexandre X. Falcão \\
  Institute of Computing\\
  University of Campinas\\
  Campinas, SP 13083-852, Brazil \\
  \texttt{afalcao@ic.unicamp.br} \\
  \And
 Pedro J. de Rezende \\
  Institute of Computing\\
  University of Campinas\\
  Campinas, SP 13083-852, Brazil \\
  \texttt{rezende@ic.unicamp.br} \\
}

\begin{document}
\maketitle
\begin{abstract}
A major challenge that prevents the training of deep learning models is the limited availability of large quantities of accurately labeled data. This shortcoming is highlighted in areas such as medical and biological sciences where data annotation becomes an expert-demanding, time-consuming, and error-prone undertaking. In this regard, semi-supervised learning tackles this challenge by capitalizing on scarce labeled and abundant unlabeled data; however, state-of-the-art methods typically depend on pre-trained features and large validation sets to learn effective representations for classification tasks. In addition, the reduced set of labeled data is often randomly sampled, neglecting the selection of more informative samples. Here, we present {\em active Deep Feature Annotation} (active-DeepFA), a method that effectively combines contrastive learning, teacher-student-based meta-pseudo-labeling and active learning to train non-pretrained CNN architectures for image classification in scenarios of scarcity of labeled and abundance of unlabeled data. It integrates deep feature annotation (DeepFA) into a co-training setup that implements two cooperative networks to mitigate confirmation bias from pseudo-labels. The method starts with a reduced set of labeled samples by warming up the networks with supervised contrastive learning. Afterward and at regular epoch intervals, label propagation is performed on the 2D projections of the networks' deep features. Next, the most reliable pseudo-labels are exchanged between networks in a cross-training fashion, while the most meaningful samples are annotated and added into the labeled set. The networks independently minimize an objective loss function comprising supervised contrastive, supervised and semi-supervised loss components, enhancing the representations towards image classification. Our approach is evaluated on three challenging biological image datasets using only 5\% of labeled samples, improving baselines and outperforming six other state-of-the-art methods. In addition, it reduces annotation effort by achieving comparable results to those of its counterparts with only 3\% of labeled data.
\end{abstract}


\section{Introduction}
\label{sec:intro}

Over the past years, deep learning methodologies have garnered significant attention and acclaim due to their remarkable versatility in addressing a range of challenges across a wide spectrum of disciplines, including computer vision, natural language processing, and speech recognition. In addition, the predictive capability of deep learning, combined with its automatic feature extraction, makes it a favored technique in medical diagnosis~\cite{dong2021survey}.

These methodologies have risen to prominence for achieving state-of-the-art results in various computer vision tasks, notably in image classification, segmentation, and object detection. This success is largely due to the vast availability of data and the simultaneous progress in both deep learning algorithms and computational power. Despite that, available data are frequently unlabeled in practical situations, presenting a significant hurdle for fully supervised methods that greatly depend on labeled data. This issue is particularly acute in fields such as medical and biological sciences, where data is very scarce and data labeling demands expertise and substantial time investment.

In light of this scenario, \emph{semi-supervised learning} (SSL) methods attempt to address the short supply of labeled data by utilizing unlabeled data to improve the generalization performance of predictive classifiers. In the field of deep neural networks, this approach is known as \emph{deep semi-supervised learning} (DSSL). Over time, a wide array of DSSL methods have emerged, showcasing various strategies and approaches~\cite{yang2023survey}. Out of these, \emph{pseudo-labeling} is renowned for its prominence due to its simple yet effective learning scheme. It essentially works by assigning pseudo-labels to unlabeled data and selecting the samples with the highest confidence predictions, which in turn are used to train and improve the model.

Recently, Benato~\etal~\cite{benato2021iterative,benato2023deep} introduced \emph{confidence Deep Feature Annotation} (conf-DeepFA), a teacher-student-based meta-pseudo-labeling method to train CNNs in semi-supervised settings. It builds on the \emph{Deep Feature Annotation} (DeepFA) technique~\cite{benato2021semi} by exploiting the non-linear 2D projections of the student CNN model's deep features for downstream label propagation within an iterative scheme. The 2D projection is achieved through t-SNE~\cite{van2014accelerating} and labels are propagated from labeled to unlabeled samples in 2D by means of a teacher graph-based classifier called OPFSemi~\cite{amorim2016improving}. In~\cite{benato2021iterative}, a threshold selection criterion is introduced to select the most confident pseudo-labeled samples, which are then utilized to train the student CNN model via cross-entropy loss. The backbone of the student CNN model comprises the VGG-16 encoder with pre-trained weights on the ImageNet dataset~\cite{russakovsky2015imagenet}. 

The conf-DeepFA method has shown to achieve high classification accuracy on several datasets from a reduced set of labeled samples ($\sim$1\% of the dataset) and without the need for a validation set, further reducing the annotation effort. Nevertheless, its implementation using non-pre-trained custom CNN architectures remains an open problem since the required amount of labeled samples to provide satisfactory results is rather high ($\sim$5\% of the dataset)~\cite{aparco2024contrastive,aparco2025consensus}. In addition, the occurrence of confirmation bias is a potential drawback for this method, where pseudo-labeling errors may have an adverse effect on the model's generalizability~\cite{arazo2020pseudo}.

More recently, Wang~\etal~\cite{wang2021sar} presented a contrastive learning and pseudo-labeling-based iterative method to classify remote sensing images in semi-supervised scenarios where labeled data are scarce and unlabeled data are abundant. The method implements two independent yet complementary CNNs that are jointly trained in a co-training fashion. The intuition behind this setup is to create two views of the training data to improve the networks' ability to learn effective feature representations. It starts by independently training the encoder section of the networks by means of the minimization of a weighted contrastive loss. Next, a classification head is added to each network for subsequent image classification. In each iteration, the networks are trained in a cross-training manner with each network playing both the roles of teacher, by producing pseudo-labels, and student, by learning from the other network's pseudo-labels. The pseudo-labels with the highest 10\% classification score are chosen to train the networks in the following iteration. An iterative categorical cross-entropy (ICE) loss function is implemented to reduce the impact of incorrect pseudo-labels during early iterations. Ultimately, an ensemble comprising the models trained across multiple iterations is used for inference on new data. This approach effectively combines contrastive learning with pseudo-labeling to achieve robust feature representation and to efficiently use unlabeled data, making it well-suited for scenarios with sparse labeled data, such as remote sensing imagery.

The work by Aparco-Cardenas~\etal~\cite{aparco2024contrastive}, hereafter called \emph{contrastive Deep Feature Annotation} (cont-DeepFA), attempts to integrate contrastive learning and deep feature annotation in a co-training fashion for the semi-supervised classification of biological images. It enables the use of non-pre-trained custom CNN architectures in the DeepFA framework by capitalizing on contrastive learning to improve the representation learning capability of two collaborative networks. Subsequently, during each iteration, the networks are trained in a teacher-student based cross-training setting, where the teacher OPFSemi performs label propagation from labeled to unlabeled samples on the non-linear 2D projection of each student network's deep features; thereafter, the pseudo-labels with the highest 10\% labeling confidence score, per class, are chosen to fine-tune the other network in a cross-training way. This has proven to effectively lessen confirmation bias and overfitting, while increasing the generalizability of the networks as iterations proceed. It was evaluated on three challenging unbalanced datasets with only 5\% of labeled samples. Despite its efficacy, the cont-DeepFA method requires a larger number of labeled samples (5\% vs 1\% of the dataset) when compared to its counterpart that uses a pre-trained CNN architecture. 

Active learning is a technique designed to identify the most informative samples from unlabeled data and present them to the oracle (\eg, a human annotator) for labeling, with the goal of minimizing labeling effort while preserving performance~\cite{ren2021survey}.

In this paper, we build on the work by Aparco-Cardenas~\etal~\cite{aparco2024contrastive} to introduce \emph{active Deep Feature Annotation} (active-DeepFA), an active contrastive-based meta-pseudo-labeling method that follows the teacher-student paradigm. It adopts a co-training strategy and capitalizes on active learning to train non-pre-trained custom CNN under conditions of short supply of labeled and abundance of unlabeled data. It implements two cooperative CNNs enhancing pseudo-labeling accuracy and mitigating confirmation bias. Moreover, it exploits active learning through the periodic labeling of the most uncertain pseudo-labeled samples during training, which in turn improves the networks' generalizability and reduces the number of required labeled samples to achieve suitable results.

The active-DeepFA method starts with a reduced set of labeled samples ($\sim$1\% of the dataset) and works by initially pre-training the networks (students) through the minimization of a weighted supervised contrastive loss for a few epochs. Afterward, a loss equal to the sum of three losses: {\it i}) a weighted supervised contrastive loss, {\it ii}) a supervised loss, and {\it iii}) a semi-supervised loss, is minimized throughout the training process for each network. The set of labeled samples is used to minimize both (i) and (ii), while a set of reliable pseudo-labels is used to minimize (iii). Label propagation is performed in intervals of a few epochs through deep feature annotation, so that the deep features of each network are projected onto 2D via t-SNE and OPFSemi (teacher) propagates labels from labeled to unlabeled samples. Next, the pseudo-labels with the highest 10\% labeling confidence score, per class, are selected to fine-tune the other network, while a small number of samples, whose pseudo-labels have the lowest labeling confidence score, are chosen to be labeled by an oracle, and then added to the set of labeled samples to minimize (i) and (ii) during subsequent epochs. The aforesaid active learning procedure halts when the number of labeled samples reaches a preset value. Lastly, an ensemble consisting of the two trained networks is used for inference on unseen data.

This method effectively combines contrastive learning, meta-pseudo-labeling and active learning to efficiently exploit unlabeled data, producing well-generalized models under conditions of scarcity of labeled and abundance of unlabeled data. Moreover, labeling the samples, identified as the most uncertain to classify, at regular epoch intervals results in improved generalization performance, since the model is continuously fed with meaningful samples during the training phase. The latter is in stark contrast to its counterparts, which randomly select the set of samples to be labeled and keep it fixed throughout the learning process.

Unlike the works by Wang~\etal~\cite{wang2021sar} and Aparco-Cardenas~\etal~\cite{aparco2024contrastive}, which use supervised contrastive learning only at the beginning of each iteration as a sort of pre-training, we incorporate it into the training loss during the whole training procedure to leverage on the samples labeled by active learning. Moreover, the training phase is reduced to a single iteration, ending up with only two networks, thus optimizing storage efficiency. 

Our approach is assessed on three challenging biological image datasets and benchmarked against a number of state-of-the-art DSSL methods to prove its effectiveness and efficiency. The datasets are unbalanced and present a high degree of resemblance among classes, making classification quite difficult.

The main contributions of this work can be summarized as:
\begin{enumerate}
    \item We present a novel active contrastive-based meta-pseudo-labeling approach that advances previous work to train non-pre-trained custom CNN architectures for image classification in conditions of short supply of labeled and abundance of unlabeled data.
    \item We integrate active learning with deep feature annotation within a co-training setting to efficiently harness the unlabeled data, which in turn improves the networks' generalizability and reduces the annotation effort by labeling the most difficult to classify samples.
\end{enumerate}

The remainder of this paper is organized as follows. Section~\ref{sec:related_work} briefly reviews the related work. Next, Section~\ref{sec:methodology} presents the proposed methodology in detail. Further, Section~\ref{sec:experiments} describes the experiments conducted and discusses the results. Lastly, Section~\ref{sec:final_remarks} concludes the paper and provides future and ongoing work directions.

\section{Related Work}
\label{sec:related_work}

Based of the different choices of architectures and unsupervised loss functions or regularization terms, DSSL methods can be divided into several categories \cite{yang2023survey}, among which pseudo-labeling, consistency regularization and hybrid methods stand out due to their straightforwardness and effectiveness. Also, deep semi-supervised methods in the literature capitalize on active learning to improve image classification while alleviating annotation burden. These methods are described in the following sub-sections.

\subsection{Pseudo-labeling methods}

Pseudo-labeling methods assign pseudo-labels to unlabeled data using only a reduced set of labeled data. In general, these can be categorized into self-training~\cite{lee2013pseudo} and disagreement-based models~\cite{zhou2008semi}. 

Self-training approaches use the model's predictions confidence to infer pseudo-labels for unlabeled data. However, recent works have proven that the model's performance can be further enhanced by using an auxiliary model (\eg, graph-based methods) in a teacher-student setup~\cite{pham2021meta,benato2021semi,benato2021iterative, benato2023deep,arazo2020pseudo}. It is noteworthy that a number of issues may arise from naive approaches, such as confirmation bias~\cite{arazo2020pseudo}, imbalance bias~\cite{wang2022debiased} or concept drift~\cite{cascante2021curriculum}, which adversely impact the generalizability of the model. Some approaches to mitigate the above issues are: to consider the pseudo-labeling confidence during the regularization process~\cite{arazo2020pseudo}, to reset the parameters of the model at the start of each learning iteration~\cite{cascante2021curriculum}, and to include the selection of the most confident pseudo-labeled samples to ensure labeling correctness~\cite{benato2021iterative,benato2023deep}. 

A well-known example in this category is \emph{Pseudo-label}~\cite{lee2013pseudo}, a self-training method that fine-tunes a pre-trained network simultaneously with both labeled and unlabeled data. The network is initially pre-trained on the labeled data in an usual supervised manner, while pseudo-labels for the unlabeled data are recomputed after each batch update and used afterwards to regularize the model.

\subsection{Consistency Regularization methods}

Consistency regularization methods build on the manifold hypothesis (data points situated on the same low-dimensional manifold should share the same label) or smoothness hypothesis (if two samples are close within the input space, their corresponding labels should match) by integrating consistency constraints into the loss function. These constraints can be adopted under different modes of perturbation (\eg, input perturbation, weights perturbation or layer perturbation) leading to a wide spectrum of methods~\cite{yang2022survey}. Well-known state-of-the-art representatives in this category are $\Pi$-Model, Mean Teacher, VAT, and UDA, which are briefly described below.

\emph{$\Pi$-Model}~\cite{laine2017temporal} forms a consensus prediction from the ensemble of predictions of instances of a single network across different training epochs, and under different regularization and input augmentation conditions. \emph{Mean Teacher}~\cite{tarvainen2017mean} works in a teacher-student setup. The teacher maintains an exponential moving average (EMA) of the student's weights over training steps, while a consistency cost minimizes the distance of the predictions of both networks. \emph{Virtual Adversarial Training} (VAT)~\cite{miyato2018virtual} uses a consistency constraint that ensures similarity between the model's outputs of the original input sample and its corresponding adversarial transformed image. \emph{Unsupervised Data Augmentation} (UDA)~\cite{xie2020unsupervised} inspects the role of noise injection in consistency training by replacing simple noising operations with advanced data augmentation methods.

\subsection{Hybrid methods}

Hybrid methods capitalize on both consistency regularization and pseudo-labeling to create more accurate and effective models in label-scarce settings. A well-known representative is \emph{FixMatch}~\cite{sohn2020fixmatch}, which assigns pseudo-labels in a self-training manner to weakly- and strongly-augmented views of an input sample, while a consistency constraint ensures their similarity.

\subsection{Active learning methods}

SSL methods leverage active learning to label the most informative and representative samples, improving model performance while minimizing labeling effort. The identification of these samples is carried out through various approaches~\cite{ren2021survey}. Some works in this category are briefly presented as follows.

Gao~\etal~\cite{gao2020consistency} propose a consistency-based semi-supervised active learning framework that works in two stages. Firstly, labeled and unlabeled data are used to train the model with cross-entropy loss and consistency-based loss, respectively. Secondly, the model outputs of unlabeled data and their augmentations are measured by a consistency-based metric. Next, unlabeled samples with low-consistency scores are selected for labeling and added to the labeled pool. Yuan~\etal~\cite{yuan2022active} propose ActiveMatch, an end-to-end SSL method that effectively combines unsupervised and supervised contrastive learning with self-training pseudo-labeling to actively select the most informative samples for labeling during the training cycle, leading to better representations towards image classification. The unlabeled samples with the lowest classification score are submitted to the oracle for labeling and then included in the labeled set. Zhang~\etal~\cite{zhang2014semi} introduce an SSL algorithm that combines co-training and active learning to improve classification performance while spending the same human annotation effort. It exploits co-training to select the most reliable samples according to the criteria of high confidence and nearest neighbor to improve the classifier, while the most informative samples are selected and sent to a human annotator for labeling.

\section{Methodology}
\label{sec:methodology}

In this section, we thoroughly present our method, outlining its constituent components and describing its execution steps.

\subsection{Notations and Definitions}
In a Semi-Supervised Learning (SSL) setting, a training set ${\cal Z}=\{\mathbf{x} | \mathbf{x} \in X\}$ comprises two disjoint subsets: a labeled set ${\cal Z}_{L}=\{(\mathbf{x}, y) | \mathbf{x} \in X, y \in Y\}$ and an unlabeled set ${\cal Z}_{U}=\{\mathbf{x} | \mathbf{x} \in X\}$, where ${\cal Z} = {\cal Z}_{L} \cup {\cal Z}_{U}$, ${\cal Z}_{L} \cap {\cal Z}_{U} = \emptyset$, and $ |{\cal Z}_{L}| \ll |{\cal Z}_{U}| $. In this notation, $ \mathbf{x} $ represents a sample, and $ y $ denotes its corresponding class label. We proceed under the assumption that the samples in both ${\cal Z}_{L}$ and ${\cal Z}_{U}$ are drawn from the same data distribution. Let us denote $B_{\cal X}$ as a set of batches for a given set ${\cal X}$.

Let $N_{1}$ and $N_{2}$ be two independent and collaborative networks that share the same architecture and structure during the training phase. Each network comprises a backbone encoder $E$ and two projection heads $P_{c}$, to perform contrastive learning, and $P_{x}$, for classification tasks. The output latent space of $E$ is used for 2D projection and label propagation.

\subsection{Network Architecture}
\label{sec:architecture}

\begin{figure}[!htb]
\begin{center}
\includegraphics[width=.7\columnwidth]{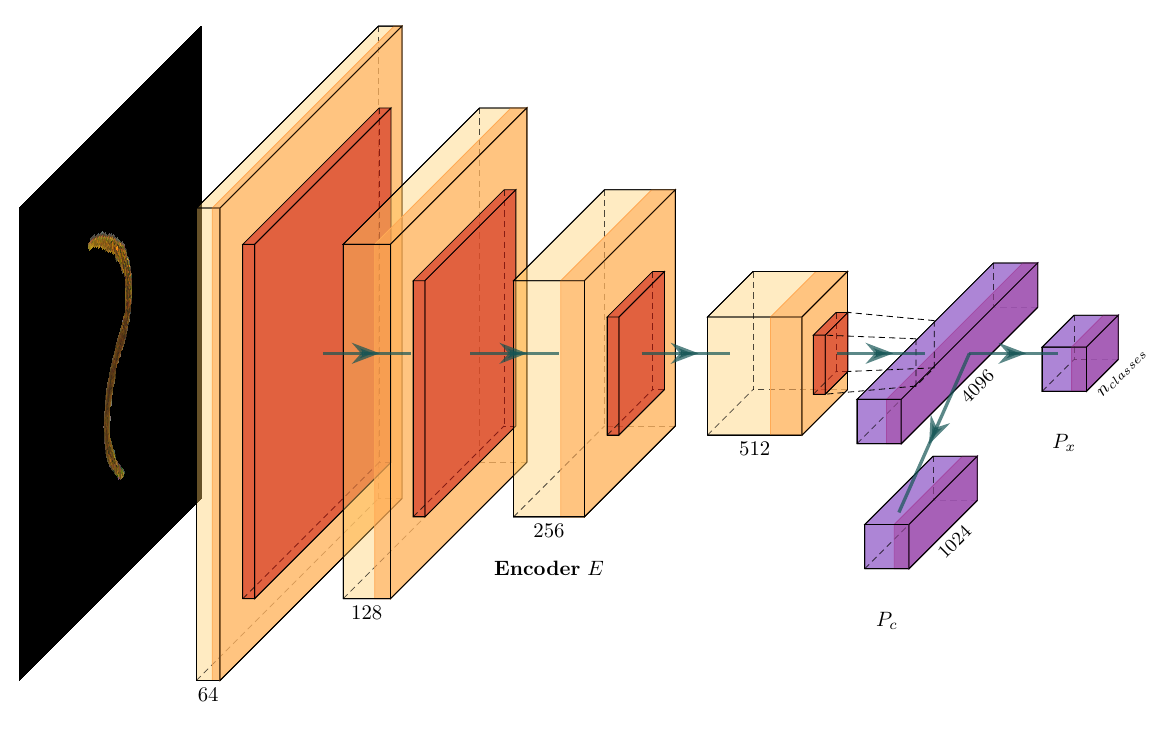}
\caption{The networks' architecture consisting of an encoder $E$ and two projection heads $P_{x}$ and $P_{c}$.}\label{fig:architecture}
\end{center}
\end{figure}

The architecture of both $N_{1}$ and $N_{2}$ comprises an \emph{encoder} $E$ consisting of four convolutional layers of 64, 128, 256, and 512 filters each, followed by a dense (fully-connected) layer of 4096 neurons. Each of the convolutional layers is constituted by a filter bank with kernel size $3\times3$, ReLU activation, max-pooling with kernel size $3\times3$ and stride of 2, and batch normalization. $P_{c}$ consists of a dense layer of 1024 neurons, while $P_{x}$ encompasses a dense layer with a number of neurons equal to the number of classes. Figure~\ref{fig:architecture} shows the networks' architecture.

\subsection{Supervised Contrastive Learning on ${\cal Z}_{L}$}

In an attempt to circumvent the encoder's pre-training prerequisite for DeepFA-based methods, we adopt supervised contrastive learning as a pre-training step during early epochs. In addition to that, it is a component of the loss function during the whole training procedure, which enhances the generalization ability of the networks. 

\begin{figure}[!htb]
\begin{center}
\includegraphics[width=.7\columnwidth]{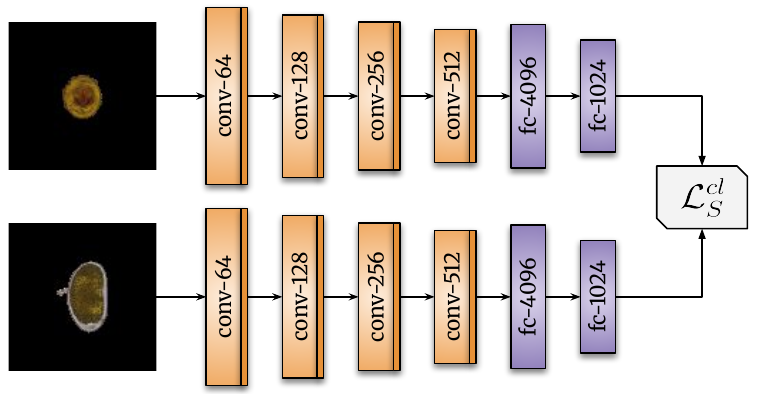}
\caption{The weight-sharing Siamese architecture adopted by each network during contrastive learning.}\label{fig:contrastive}
\end{center}
\end{figure}

In this regard, contrastive learning acts as a dimensionality reduction technique by mapping, in a contrastive manner, a set of high-dimensional data points (\eg, images) to representations in a lower-dimensional space, such that representations of semantically similar pairs are drawn closer, whereas those of dissimilar pairs are pushed away from each other in the lower-dimensional space. During the learning phase, each network adopts a Siamese architecture comprising two weight-sharing subnetworks as shown in Figure~\ref{fig:contrastive}. Let $\mathbf{x}_{1}, \mathbf{x}_{2}\in{\cal Z}_{L}$ be a pair of labeled input images, and let $\Tilde{\mathbf{x}}_{1}, \Tilde{\mathbf{x}}_{2}$ be their respective augmented versions. Over the course of training, $\Tilde{\mathbf{x}}_{1}$ and $\Tilde{\mathbf{x}}_{2}$ serve as input to the different branches of the Siamese network. The distance function ${\cal D}$ between the lower-dimensional representations of $\Tilde{\mathbf{x}}_{1}$ and $\Tilde{\mathbf{x}}_{2}$ is defined as the Euclidean distance:

\begin{equation}
    {\cal D}(\Tilde{\mathbf{x}}_{1}, \Tilde{\mathbf{x}}_{2}) = \|P_{c}(E(\Tilde{\mathbf{x}}_{1})) - P_{c}(E(\Tilde{\mathbf{x}}_{2}))\|
\end{equation}

Let $y_{t}$ be a binary label assigned to the pair $\mathbf{x}_{1},\mathbf{x}_{2}$, where $y_{t}=0$ if they are deemed similar and $y_{t}=1$ otherwise. To compensate for the large difference between the number of similar and dissimilar pairs, a weight factor $\tau$ is incorporated into the contrastive loss function. The weighted supervised contrastive loss function is defined as

\begin{equation}
    {\cal L}_{S}^{cl} = \sum_{i=1}^{|S|}L(p^{i})
\end{equation}

\begin{equation}
    L(p^{i}) = \tau(1 - y_{t})\frac{1}{2}({\cal D}^{i})^{2}+(y_{t})\frac{1}{2}\{\max(0, m - {\cal D}^{i})\}^{2}
\end{equation}

\noindent where $p^{i}=(y_{t},\Tilde{\mathbf{x}}_{1},\Tilde{\mathbf{x}}_{2})^{i}$ is the $i$-th labeled sample pair and $S$ is the set of possible sample pairs for a given batch $B^{l}\in B_{{\cal Z}_{L}}$, such that $|S|=\binom{|B^{l}|}{2}$. The value of the margin $m$ is empirically set to $2$. 

Let $m$ be the number of classes of the dataset and let $k$ be the size of each class for a given balanced dataset. The ratio between the number of similar and dissimilar pairs can be computed as

\begin{equation}
    \frac{m\binom{k}{2}}{\binom{mk}{2} - m\binom{k}{2}} = \frac{k-1}{k(m-1)} \approx \frac{1}{m-1}.
\end{equation}

Therefore, the value of $\tau$ is set to $m - 1$ to compensate for the contribution of similar pairs in ${\cal L}_{S}^{cl}$ as suggested in~\cite{wang2021sar}.

\subsection{Supervised Learning on ${\cal Z}_{L}$}

For a given batch $B^{l}\in B_{{\cal Z}_{L}}$, the supervised loss function is defined as

\begin{equation}
    {\cal L}_{S} = \frac{1}{|B^{l}|} \sum_{i=1}^{|B^{l}|}CE(y_{i},\mathbf{q}_{i})
\end{equation}

\noindent where $y_{i}$ and $\mathbf{q}_{i}$ are the label and the logit prediction for the $i$-th sample, and $CE(\cdot,\cdot)$ represents the cross-entropy loss.

\subsection{Semi-Supervised Learning on ${\cal Z}_{U}$}

The pseudo-labeling procedure is carried out via label propagation by means of OPFSemi~\cite{amorim2016improving}, a graph-based technique inspired on the optimum-path forest (OPF) methodology, which has been adopted in other recent works for the same purpose~\cite{benato2021semi,benato2021iterative,benato2023deep}.

Let $\phi: \mathrm{R}^{n}\to\mathrm{R}^{2}$ be a function that non-linearly maps the latent space output of the encoder $E$ into 2D. For each network, $\phi({\cal Z})$ is computed by forward passing ${\cal Z}$ through $E$ and projecting the output deep features into 2D via t-SNE. The OPFSemi technique is executed on the graph induced by the samples in the projection set $\phi({\cal Z})$. It starts by modeling $\phi({\cal Z})$ as a complete graph, where samples are encoded as nodes. The projection subset of labeled samples $\phi({\cal Z}_{L})$ acts as the set of prototypes and their labels are propagated to their most closely connected unlabeled samples in $\phi({\cal Z}_{U})$, resulting in a graph partitioned into optimum-path trees rooted at $\phi({\cal Z}_{L})$.

Let $\lambda'(u)$ be the pseudo-label assigned to $u\in\phi({\cal Z}_{U})$ via OPFSemi's label propagation, and let $\lambda(u)$ be the ground-truth label of $s\in\phi({\cal Z}_{L})$, such that $s$ is the root of the optimum-path tree to which $u$ is connected -- \ie, $\lambda'(u)=\lambda(s)$. Let $c(u)$ be the cost of the optimum-path offered from $s$ to $u$, and let $c'$ be the cost of the second smallest optimum-path offered to $u$ from $t\in\phi({\cal Z}_{L})$, where $s\neq t$ and $c'>c(u)$. At the end of the execution, a labeling confidence value defined as

\begin{equation}
    v(u) = \frac{c'}{c(u)+c'}\in[0,1]
\end{equation}

\noindent is calculated and assigned to each $u\in\phi({\cal Z}_{L})$.

Afterward, we pick the pseudo-labels with the highest 10\% labeling confidence value, per class. This procedure guarantees the selection of reliable pseudo-labels that encompass all classes in the dataset. In contrast, using a fixed labeling confidence threshold may fail to select pseudo-labels from classes with a low mean labeling confidence value.

Let $Y'$ be the set of pseudo-labels with the highest 10\% labeling confidence value, per class. We define the set ${\cal Z}_{U}^{pl} = \{(\mathbf{x}, y')\,|\,\mathbf{x}\in{\cal Z}_{U}, y'\in Y'\}$, where $y'$ is the pseudo-label assigned to sample $\mathbf{x}$. For a given batch $B^{pl}\in B_{{\cal Z}_{U}^{pl}}$, the semi-supervised loss function is defined as

\begin{equation}
    {\cal L}_{U}^{ssl} = \frac{1}{|B^{pl}|} \sum_{i=1}^{|B^{pl}|}CE(y'_{i},\mathbf{q_{i}})
\end{equation}

\noindent where $y'_{i}$ and $\mathbf{q}_{i}$ are the pseudo-label and the logit prediction for the $i$-th sample. During training, we apply strong augmentations to the samples of ${\cal Z}_{U}^{pl}$, as defined in~\cite{sohn2020fixmatch}, to enhance the learning performance and the generalizability of the network. 

\subsection{Active Learning and Cross-Training}

Once label propagation is concluded, we identify the most uncertain samples as the ones whose pseudo-labels have the lowest labeling confidence value, which are the ones located at the decision boundary of the optimum-path trees. These samples are regarded as the most difficult to classify and thus assumed to contain meaningful information to improve generalization performance. Let ${\cal Z}_{A}^{(i)}$ be the set of samples chosen by active learning for $N_{i},\,i\in\{1, 2\}$. The set ${\cal Z}_{A}^{(i)}$ comprises the $k_{\text{active}}$ samples whose pseudo-labels have the lowest labeling confidence values, for a fixed value of $k_{\text{active}}$. These sets are then joined and the $k_{\text{active}}$ samples with the lowest labeling confidence values are sent to the oracle to be labeled and included in the labeled set ${\cal Z}_{L}$ to minimize both ${\cal L}_{S}^{cl}$ and ${\cal L}_{S}$ during the following epochs. This procedure is halted when a preset maximum number of samples labeled by active learning $n_{\text{active}}$ is achieved. 

The conventional approach for the pseudo-labeling training stage is to re-train the network with the pseudo-labels generated by itself. Nevertheless, this may inadvertently provoke overfitting and confirmation bias, thus hampering network learning. In this respect, the idea of using the pseudo-labels generated by another model stems from the notion that the generalization capability of the network is enhanced when exposed to new information. Let ${\cal Z}_{U}^{pl(i)}$ be the set of picked pseudo-labeled samples for $N_{i},\,i\in\{1, 2\}$. Therefore, after each pseudo-labeling stage, the set ${\cal Z}_{U}^{pl(2)}$ is used to minimize ${\cal L}_{U}^{ssl}$ for network $N_{1}$ and vice versa for $N_{2}$, performing this in a cross-training manner. Ultimately, the objective loss function that each network minimizes during the training phase is defined as

\begin{equation}
    {\cal L} = {\cal L}_{S}^{cl} + {\cal L}_{S} + {\cal L}_{U}^{ssl}.
\end{equation}

After the training process has been completed, we use both $N_{1}$ and $N_{2}$ in an ensemble fashion to predict new data. Let $\vec{pv}^{i}(u)$ the predicted probability vector of $N_{i}$ for sample $\mathbf{x}$. The label of $\mathbf{x}$, denoted as $\lambda(\mathbf{x})$, is computed as the index of the highest value of the average of the predicted probability vectors of $N_{1}$ and $N_{2}$ in the following way

\begin{equation}
    \lambda(\mathbf{x)} = \argmax\left\{\frac{\vec{pv}^{1}(\mathbf{x})+\vec{pv}^{2}(\mathbf{x})}{2}\right\}.
\end{equation}

\begin{figure*}[!htb]
\begin{center}
\includegraphics[width=\columnwidth]{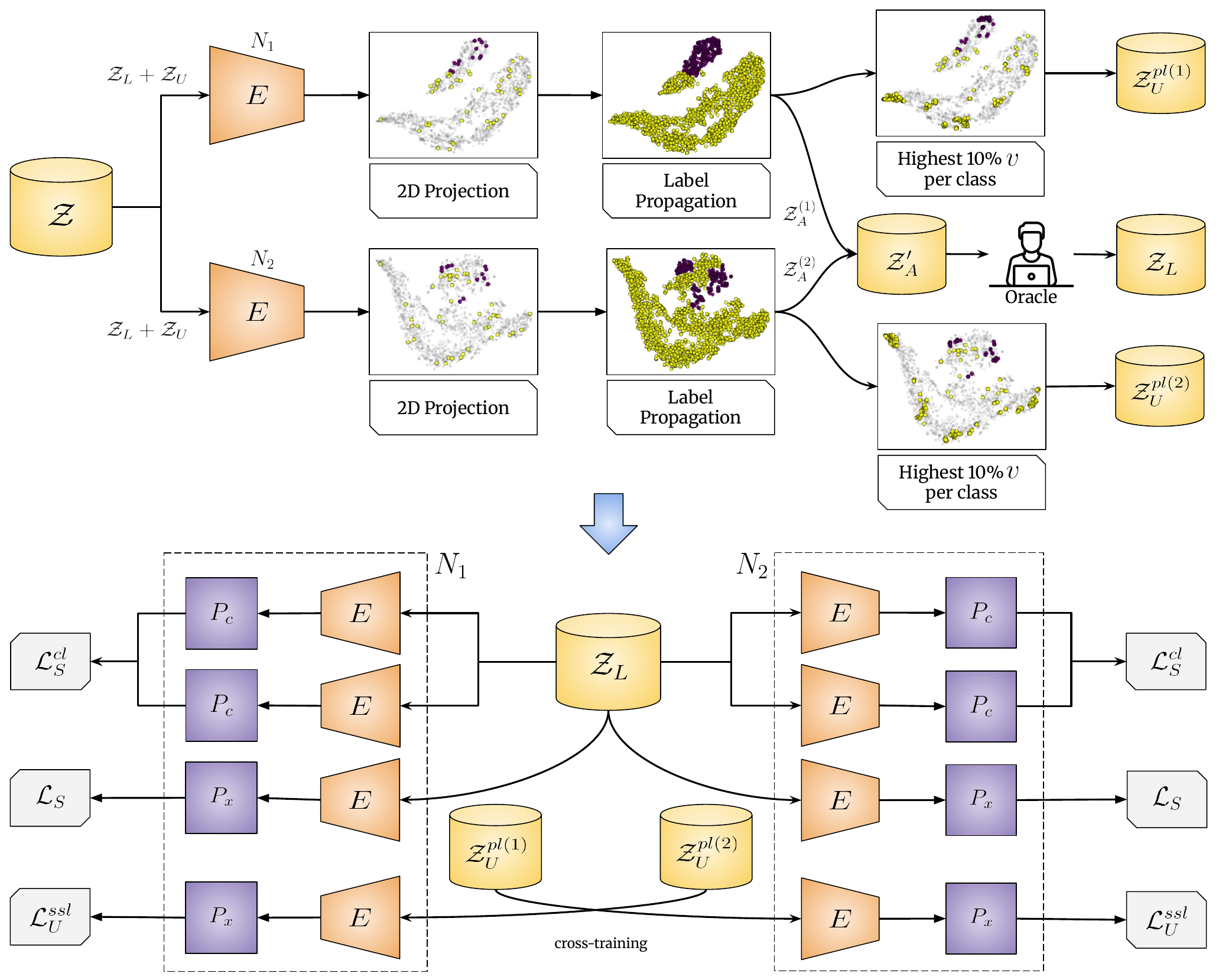}
\caption{Overview of the active-DeepFA method. For each network, the output deep features of $E({\cal Z})$ are projected into 2D using t-SNE and labeling propagation is executed by OPFSemi on the 2D space. Next, the sets ${\cal Z}_{U}^{pl(i)},{{\cal Z}_{A}^{(i)}}$ are constructed for $N_{i},\,i\in\{1,2\}$. The set ${\cal Z}'_{A}$ is created by joining the sets ${\cal Z}_{A}^{(1)},{\cal Z}_{A}^{(2)}$ and picking the $k_{\text{active}}$ samples whose pseudo-labels have the lowest $v$ value. Then, the elements of ${\cal Z}'_{A}$ are sent to the oracle to be labeled and added to ${\cal Z}_{L}$. Afterward, the losses ${\cal L}_{S}^{cl},{\cal L}_{S}$ are computed on ${\cal Z}_{L}$, while ${\cal L}_{U}^{ssl}$ is calculated on ${\cal Z}_{U}^{pl(2)}$ for $N_{1}$ and vice versa for $N_{2}$ in a cross-training manner.}\label{fig:activedeepfa} 
\end{center}
\end{figure*}

\subsection{Overview of the Proposed Method}

An overview of the procedural steps of active-DeepFA is outlined in Algorithm~\ref{alg:algorithm}. The method starts with two networks $N_{1},N_{2}$ (students) sharing the same architecture and structure. Since we desire that each network produce a unique representation for a same input image, we begin by independently conducting Kaiming He initialization~\cite{he2015delving} on the weights of each network (Line~\ref{line:kaiminghe}). We also keep a counter $c_{\text{active}}$ of the number of samples labeled by active learning, which is initially set to zero (Line~\ref{line:cactive}). The training process is executed for a preset number of epochs, $n_{\text{epochs}}$. As a form of pre-training, we perform contrastive learning by minimizing ${\cal L}_{S}^{cl}$ on ${\cal Z}_{L}$ for a fixed number of warming epochs, $w_{\text{epochs}}$, where $w_{\text{epochs}} < n_{\text{epochs}}$ (Lines~\ref{line:initwarm}-\ref{line:endwarm}).

Afterward, for the epoch immediately after the warming step -- \ie, at epoch $w_{\text{epochs}}+1$ -- and at intervals of $e_{\text{int}}$ epochs -- \ie, at every epoch divisible by $e_{\text{int}}$, we project the output deep features of $E({\cal Z})$ onto 2D using t-SNE and execute label propagation from $\phi({\cal Z}_{L})$ to $\phi({\cal Z}_{U})$ on the 2D space via OPFSemi (teacher). Next, the sets ${\cal Z}_{U}^{pl}$ and ${\cal Z}_{A}$ are constructed exploiting the labeling confidence values assigned to the pseudo-labels (Lines~\ref{line:initprop}-\ref{line:endprop}). Let ${\cal Z}_{U}^{pl(1)},\,{\cal Z}_{A}^{(1)}$ and ${\cal Z}_{U}^{pl(2)},\,{\cal Z}_{A}^{(2)}$ be the sets obtained from $N_{1}$ and $N_{2}$, respectively. If $c_{\text{active}}$ is less than a fixed maximum number $n_{\text{active}}$, we proceed to construct the set ${\cal Z}'_{A}$ by joining the sets ${\cal Z}_{A}^{(1)},{\cal Z}_{A}^{(2)}$ and selecting the $k_{\text{active}}$ samples with the lowest labeling confidence value. The samples of ${\cal Z}'_{A}$ are submitted to the oracle to be labeled and subsequently added to ${\cal Z}_{L}$. Next, the value of $c_{\text{active}}$ is updated accordingly (Lines~\ref{line:initactive}-\ref{line:endactive}). 

Then, for each network, we compute the losses ${\cal L}_{S}^{cl}, {\cal L}_{S}$ on ${\cal Z}_{L}$, and ${\cal L}_{U}^{ssl}$ on the strong augmented samples of ${\cal Z}_{U}^{pl(2)}$ for $N_{1}$ and vice versa for $N_{2}$. The losses are subsequently added and the network is updated accordingly to minimize this sum (Lines~\ref{line:initssl}-\ref{line:endssl}). Lastly, the trained networks $N_{1},N_{2}$ are used in an ensemble manner to predict on new data.

\begin{algorithm}[!hbt]
\caption{active-DeepFA algorithm}\label{alg:algorithm}
\hspace*{\algorithmicindent} \textbf{Input:} ${\cal Z}$, $N_{1}$, $N_{2}$, $n_{\text{epochs}}$, $w_{\text{epochs}}$, $e_{\text{int}}$, $n_{\text{active}}$, $k_{\text{active}}$\\
 \hspace*{\algorithmicindent} \textbf{Output:} Trained networks $N_{1},N_{2}$
\begin{algorithmic}[1]
  \State Initialize $N_{1},N_{2}$ with Kaiming He initialization\label{line:kaiminghe}
  \State $c_{\text{active}}\gets 0$\label{line:cactive}
  \For{$e\gets 1$ to $n_{\text{epochs}}$}
    \If{$e\leq w_{\text{epochs}}$}\label{line:initwarm}
      \For{$B^{l}\in B_{{\cal Z}_{L}}$}
        \For{$N\in\{N_{1},N_{2}\}$}
          \State Compute ${\cal L}_{S}^{cl}$ using $B^{l}$ 
          \State ${\cal L}\gets{\cal L}_{S}^{cl}$
          \State Update $N$ to minimize ${\cal L}$\label{line:endwarm}
        \EndFor
      \EndFor
    \Else
      \If{$e\,\%\,e_{\text{int}} = 0$ {\bf or} $e = w_{\text{epochs}} + 1$}\label{line:initprop}
        \For{$N\in\{N_{1},N_{2}\}$}
          \State Project $E({\cal Z})$ onto 2D
          \State Propagate labels from $\phi({\cal Z}_{L})$ to $\phi({\cal Z}_{U})$
          \State Construct the sets ${\cal Z}_{U}^{pl}$ and ${\cal Z}_{A}$\label{line:endprop}
        \EndFor
        \If{$c_{active} < n_{active}$}\label{line:initactive}
          \State Construct ${\cal Z}'_{A}$ from ${\cal Z}_{A}^{(1)}$ and ${\cal Z}_{A}^{(2)}$
          \State Sent ${\cal Z}'_{A}$ to the oracle for labeling
          \State ${\cal Z}_{L}\gets{\cal Z}_{L}\cup{\cal Z}'_{A}$
          \State $c_{\text{active}}\gets c_{\text{active}} + k_{\text{active}}$\label{line:endactive}
        \EndIf
      \EndIf
      \For{$B^{l}\in B_{{\cal Z}_{L}}, B^{pl}\in B_{{\cal Z}_{U}^{pl}}$}\label{line:initssl}
        \For{$N\in\{N_{1},N_{2}\}$}
          \State Compute ${\cal L}_{S}^{cl},{\cal L}_{S}$ using $B^{l}$
          \State Compute ${\cal L}_{U}^{ssl}$ using $B^{pl}$ \Comment{cross-training}
          \State ${\cal L}\gets{\cal L}_{S}^{cl}+{\cal L}_{S}+{\cal L}_{U}^{ssl}$
          \State Update $N$ to minimize ${\cal L}$\label{line:endssl}
        \EndFor
      \EndFor
    \EndIf
  \EndFor
  \State {\bf return} $N_{1},N_{2}$
\end{algorithmic}
\end{algorithm}

\section{Experiments and Results}
\label{sec:experiments}

This section evaluates the performance of active-DeepFA on three challenging datasets, compares it to the state-of-the-art methods, and provides an analysis and discussion of the experimental results.

\subsection{Datasets}

\begin{figure}[!hbt]
\centering
\subfloat{
\includegraphics[height=0.1\textheight]{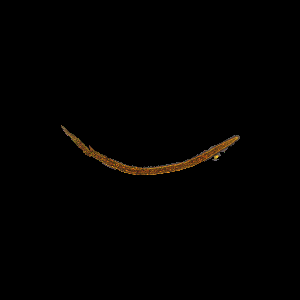}
\includegraphics[height=0.1\textheight]{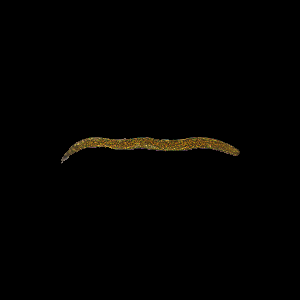}}
\subfloat{
\includegraphics[height=0.1\textheight]{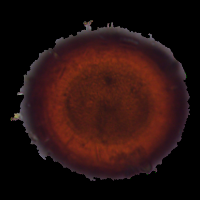}
\includegraphics[height=0.1\textheight]{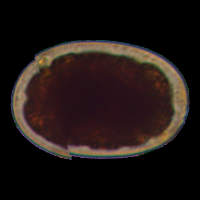}}
\subfloat{
\includegraphics[height=0.1\textheight]{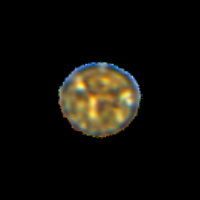}
\includegraphics[height=0.1\textheight]{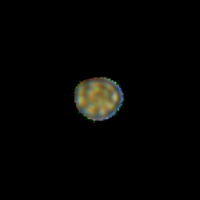}
}
\\
\vspace{-3mm}
\setcounter{subfigure}{0}
\hspace{-1.5mm}
\subfloat[{\it Helminth larvae}]{
\includegraphics[height=0.1\textheight]{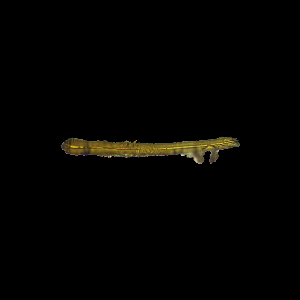}
\includegraphics[height=0.1\textheight]{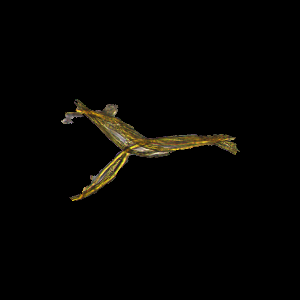}}
\subfloat[{\it Helminth eggs}]{
\includegraphics[height=0.1\textheight]{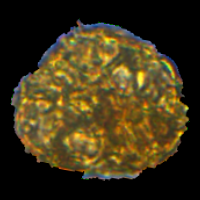} 
\includegraphics[height=0.1\textheight]{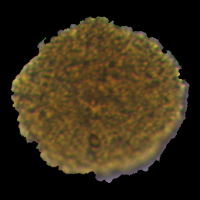}}
\subfloat[{\it Protozoan cysts}]{
\includegraphics[height=0.1\textheight]{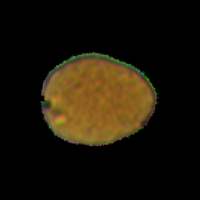} 
\includegraphics[height=0.1\textheight]{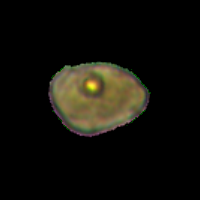}}
\caption{Images from the \textit{Helminth larvae}, \textit{Helminth eggs} and \textit{Protozoan cysts} datasets. The first row shows images of parasites, while the second row exhibits images of impurities.}
\label{fig:datasets}
\end{figure}

We assess the active-DeepFA method on three demanding datasets from a parasite biological image collection~\cite{suzuki2013automated}. The collection consists of optical microscopy images of the most common human intestinal parasites in Brazil. These parasites are widespread in countries with tropical, subtropical, and equatorial climates, posing substantial public health challenges. The images are rescaled to $200\times200\times3$ pixels. The datasets comprising this collection are listed as follows: {\it i})~\emph{Helminth larvae}; {\it ii})~\emph{Helminth eggs}; and {\it iii})~\emph{Protozoan cysts}. These datasets are characterized by their class imbalance and the presence of similar features among distinct classes, which increases the difficulty of network learning. Table~\ref{tab:datasets} shows a brief description of the datasets, while some images of parasites and impurities for ({\it i})--({\it iii}) are displayed in Figure~\ref{fig:datasets}.

\begin{table}[!hbt]
    \caption{Description of the datasets.}
    \centering
    \renewcommand{\arraystretch}{1.}
    \resizebox{.7\columnwidth}{!}{%
    \begin{tabular}{l | r | l | c}
        {\bf Dataset} & {\bf Number} & {\bf Category} & {\bf Class ID} \\
        \hline
        \hline
        \multirow{3}{*}{\it Helminth larvae}
        & 446 & {\it Strongyloides stercoralis} & 1 \\
        & 3 068 & Impurities & 2 \\
        & \textbff{3 514} & \textbff{Total} & \\
        \hline
        \multirow{10}{*}{\it Helminth eggs}
        & 348 & {\it Hymenolepis nana} & 1 \\
        & 80 & {\it Hymenolepis diminuta} & 2 \\
        & 148 & Ancylostomatidae & 3 \\
        & 122 & {\it Enterobius vermicularis} & 4 \\
        & 337 & {\it Ascaris lumbricoides} & 5 \\
        & 375 & {\it Trichuris trichiura} & 6 \\
        & 122 & {\it Schistosoma mansoni} & 7 \\
        & 236 & {\it Taenia} spp. & 8 \\
        & 3 344 & Impurities & 9 \\
        & \textbff{5 112} & \textbff{Total} & \\
        \hline
        \multirow{8}{*}{\it Protozoan cysts}
        & 719 & {\it Entamoeba coli} & 1 \\
        & 78 & {\it Entamoeba histlytica / E. dispar} & 2 \\
        & 724 & {\it Endolimax nana} & 3 \\
        & 641 & {\it Giardia duodenalis} & 4 \\
        & 1 501 & {\it Iodamoeba bütschlii} & 5 \\
        & 189 & {\it Blastocystis hominis} & 6 \\
        & 5 716 & Impurities & 7 \\
        & {\bf 9 568} & \textbf{Total} & \\
        \hline
    \end{tabular} }
    \label{tab:datasets}
\end{table}

\subsection{Dataset Preparation}
\label{sec:dataset_preparation}

We split the dataset ${\cal Z}_{D}$ into $\frac{2}{3}$ for the training set ${\cal Z}$ and the remaining $\frac{1}{3}$ for the test set ${\cal Z}_{T}$ (${\cal Z}_{D}={\cal Z}\cup{\cal Z}_{T}$) by means of three-fold cross-validation -- creating 3 train/test splits. A validation set is omitted, as its inclusion would go against the purpose of this work by increasing the annotation effort.

We emulate the conditions of short supply of labeled data by partitioning the training set for each split into labeled ${\cal Z}_{L}$ and unlabeled ${\cal Z}_{U}$ sets (${\cal Z}={\cal Z}_{L}\cup{\cal Z}_{U}$), such that $|{\cal Z}_{L}|=1\%\cdot|{\cal Z}_{D}|$ and $|{\cal Z}_{U}|=65.66\%\cdot|{\cal Z}_{D}|$. The $1\%$ of labeled samples are randomly picked from ${\cal Z}$ for each split in a stratified manner. Table~\ref{tab:lab_unlab} shows the number of labeled and unlabeled samples for each dataset.

\begin{table}[!htb]
    \caption{Number of labeled and unlabeled samples for each dataset.}
    \centering
    \renewcommand{\arraystretch}{1.25}
    \resizebox{.35\columnwidth}{!}{%
    \begin{tabular}{c | c | c | c}
        & {\bf H. larvae} & {\bf H. eggs} & {\bf P. cysts} \\
        \hline
        \hline
        {$|{\cal Z}_{L}|$} & 36 & 56 & 100 \\
        \hline
        {$|{\cal Z}_{U}|$} & 176 & 3356 & 6281 \\
        \hline
    \end{tabular} }
    \label{tab:lab_unlab}
\end{table}

\subsection{Experimental Setup}

All implementations were carried out in PyTorch 1.8.0. For t-SNE, we used the implementation from the Multicore t-SNE library~\cite{ulyanov2016multicore} with perplexity value of 50. The OPFSemi algorithm is parameter-free, so no tuning was required. As for the optimization hyperparameters, we used stochastic gradient descent (SGD) as optimizer with batch size value of 32, learning rate value of $3\cdot10^{-4}$, momentum value of $0.9$, weight decay value of $5\cdot10^{-4}$ and Nesterov momentum. The learning rate adopts a cosine learning rate decay~\cite{loshchilov2017sgdr}, whose value at step $t$ is $lr = lr_{0} \cdot cos(\frac{7\pi t}{16T})$, where $lr_{0}$ is the initial learning rate value and $T$ is the total number of training steps. For the sake of comparison purposes, we use the results from~\cite{aparco2024contrastive}, since the same three-fold cross-validation splits are employed and the networks have an almost identical architecture. 

The only difference lies on the projection head, where the architecture presented in~\cite{aparco2024contrastive} has a single projection head consisting of two hidden dense (fully-connected) layers of 512 and 256 neurons each. The architecture used for our method is presented in Section~\ref{sec:architecture}.

The active-DeepFA method starts with $|{\cal Z}_{L}|=1\%\cdot|{\cal Z}_{D}|$. We use a number of warming epochs equal to $w_{\text{epochs}}=15$. The warming-up phase using supervised contrastive learning allows the network to learn meaningful representations for both pseudo-labeling and active learning tasks. The values of the number of epochs per interval and the number of samples labeled by active learning per interval are empirically set to $e_{\text{int}}=5$ and $k_{\text{active}}=2$. The number of epochs $n_{\text{epochs}}$ varies in function of the maximum number of labeled samples by active learning $n_{\text{active}}$. Since our aim is to make a comparison with the results in~\cite{aparco2024contrastive}, we continue labeling samples via active learning until we reach $|{\cal Z}_{L}|=5\%\cdot|{\cal Z}_{D}|$. In this regard, the values of $n_{\text{epochs}}$ and $n_{\text{active}}$ for each dataset are shown in Table~\ref{tab:epochs_active}.

\begin{table}[!htb]
    \caption{Values of $n_{\text{epochs}}$ and $k_{\text{active}}$ for each dataset.}
    \centering
    \renewcommand{\arraystretch}{1.25}
    \resizebox{.35\columnwidth}{!}{%
    \begin{tabular}{c | c | c | c}
        & {\bf H. larvae} & {\bf H. eggs} & {\bf P. cysts} \\
        \hline
        \hline
        {$n_{\text{epochs}}$} & 375 & 535 & 983 \\
        \hline
        {$n_{\text{active}}$} & 140 & 200 & 379 \\
        \hline
    \end{tabular} }
    \label{tab:epochs_active}
\end{table}

We use accuracy and Cohen's $\kappa$, hereafter denoted as $\kappa$, to evaluate the performance of the method. The $\kappa$ metric, where $\kappa\in [-1, 1]$, provides a more reliable measure than accuracy for unbalanced datasets. It quantifies the degree of disagreement between the classifier's prediction and the ground-truth, where $\kappa = 1$ and $\kappa = -1$ indicate full agreement and disagreement, respectively. For each method, the mean and standard deviation of both accuracy and $\kappa$ across splits are presented.

\subsection{Comparison with baselines}

In this experiment, we intend to benchmark our method against its direct baselines, from which this work derives. The main baselines of our approach are the method by Wang~\etal~\cite{wang2021sar}, conf-DeepFA~\cite{benato2023deep} and cont-DeepFA~\cite{aparco2024contrastive}. It is worth noting that the active-DeepFA method and its baselines do not utilize a validation set. The experiment configuration and hyperparameter values for all baselines can be found in~\cite{aparco2024contrastive}. It is important to highlight that all the baselines start with $|{\cal Z}_{L}|=5\%\cdot|{\cal Z}_{D}|$, while active-DeepFA starts with only $|{\cal Z}_{L}|=1\%\cdot|{\cal Z}_{D}|$ and expands ${\cal Z}_{L}$ through active learning to reach $|{\cal Z}_{L}|=5\%\cdot|{\cal Z}_{D}|$, thus all end up using the same amount of labeled data.

The mean and standard deviation of both accuracy and $\kappa$ across splits are shown in Table~\ref{tab:baselines}. It can be observed that active-DeepFA achieves better generalization on the test set for all three datasets.

\begin{table}[!htb]
    \caption{Test results of accuracy and $\kappa$ for baselines with $|{\cal Z}_{L}|=5\%\cdot|{\cal Z}_{D}|$.}
    \centering
    \renewcommand{\arraystretch}{1.2}
    \resizebox{.7\columnwidth}{!}{%
    \begin{tabular}{ l | c | c | c | c}
        \multirow{2}{*}{\bf Method} & \multirow{2}{*}{\bf Metric} & \multicolumn{3}{c}{\bf Datasets} \\
        \cline{3-5}
        & & {\it Helminth larvae} & {\it Helminth eggs} & {\it Protozoan cysts} \\
        \hline
        \multirow{2}{*}{Wang~\etal~\cite{wang2021sar}} & accuracy & 0.970 $\pm$ 0.022 & 0.906 $\pm$ 0.004 & 0.886 $\pm$ 0.013 \\
        & $\kappa$ & 0.878 $\pm$ 0.078 & 0.822 $\pm$ 0.006 & 0.806 $\pm$ 0.021 \\
        \hline
        \multirow{2}{*}{conf-DeepFA~\cite{benato2023deep}} & accuracy & 0.966 $\pm$ 0.016 & 0.884 $\pm$ 0.010 & 0.799 $\pm$ 0.060 \\
        & $\kappa$ & 0.867 $\pm$ 0.064 & 0.785 $\pm$ 0.024 & 0.663 $\pm$ 0.086 \\
        \hline
        \multirow{2}{*}{cont-DeepFA~\cite{aparco2024contrastive}} & accuracy & 0.976 $\pm$ 0.015 & 0.946 $\pm$ 0.005 & 0.922 $\pm$ 0.009 \\
        & $\kappa$ & 0.889 $\pm$ 0.064 & 0.901 $\pm$ 0.009 & 0.870 $\pm$ 0.014 \\
        \hline
        \multirow{2}{*}{active-DeepFA} & accuracy & \textbff{0.992 $\pm$ 0.003} & \textbff{0.965 $\pm$ 0.004} & \textbff{0.962 $\pm$ 0.009} \\
        & $\kappa$ & \textbff{0.959 $\pm$ 0.013} & \textbff{0.937 $\pm$ 0.007} & \textbff{0.937 $\pm$ 0.015} \\
        \hline
    \end{tabular} }
    \label{tab:baselines}
\end{table}

\subsubsection{Influence of supervised contrastive learning}

The results suggest that supervised contrastive learning acts as an effective weight initializer for DeepFA-based methods. In this regard, we circumvent the requirement for pre-trained encoders while reducing annotation costs by learning from a reduced set of labeled samples, thus enabling the use of non-pretrained custom CNN architectures for DeepFA-based methods as conf-DeepFA.

In stark contrast to the baselines that uniquely use supervised contrastive learning as a sort of pre-training step, we further exploit it by incorporating it into the objective loss function throughout the whole training process. This procedure is enhanced by the dynamic labeled set, which is periodically expanded by active learning at regular epoch intervals. These informative samples incrementally improve the low-dimensional feature representation learned by the network from the weighted supervised contrastive loss. Moreover, the implementation of an additional projection head allows the simultaneous minimization of the three losses -- supervised, semi-supervised and contrastive -- thereby, boosting the representation ability of the network as demonstrated in the results. In contrast, the utilization of a single projection head for both contrastive and classification tasks may interfere with each other's learning by overwriting the previously updated weights.

The work by Wang~\etal~\cite{wang2021sar} performs contrastive learning directly on the output representations of the last convolutional layer, bypassing the use of a projection head. In this scenario, the Euclidean distance is applied to a high-dimensional space, making it susceptible to the curse of dimensionality, which adversely impacts the quality of learned representations. On the other hand, the approach by Aparco-Cardenas~\etal~\cite{aparco2024contrastive} employ a projection head to perform contrastive learning, afterward a decision layer is attached on top of it for downstream classification tasks. It should be noted that this strategy may not be suitable for the concomitant learning of both contrastive and classification tasks due to weight overwriting issues.
 
\subsubsection{Influence of active learning}

\begin{figure*}[!htb]
\begin{center}
\includegraphics[width=\columnwidth]{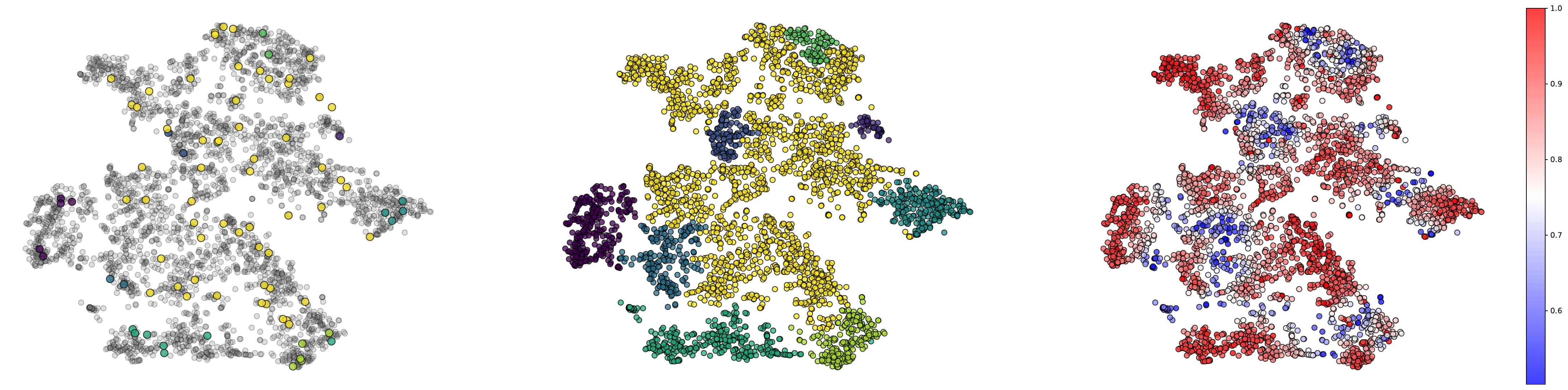}
\caption{Heatmap of the labeling confidence values after labeling propagation on the 2D projection of the {\em Helminth eggs}'s deep features via OPFSemi. Left: the prototypes from $\phi({\cal Z}_{L})$ are non-gray-colored with a different color per class, while samples from $\phi({\cal Z}_{U})$ are gray-colored. Middle: result of label propagation from $\phi({\cal Z}_{L})$ to $\phi({\cal Z}_{U})$. Right: heatmap of the labeling confidence values, where blue-colored samples with the lowest labeling confidence values lie close to decision boundary regions.}\label{fig:certainty} 
\end{center}
\end{figure*}

\begin{figure*}[!htb]
\centering
\subfloat[{\em Helminth larvae}]{\includegraphics[width=0.33\textwidth]{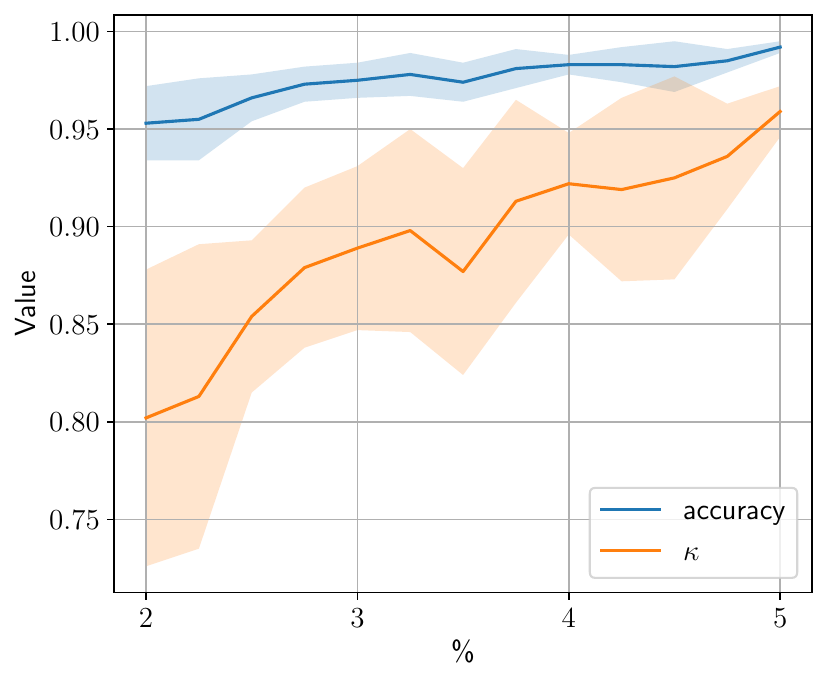}}
\subfloat[{\em Helminth eggs}]{\includegraphics[width=0.33\textwidth]{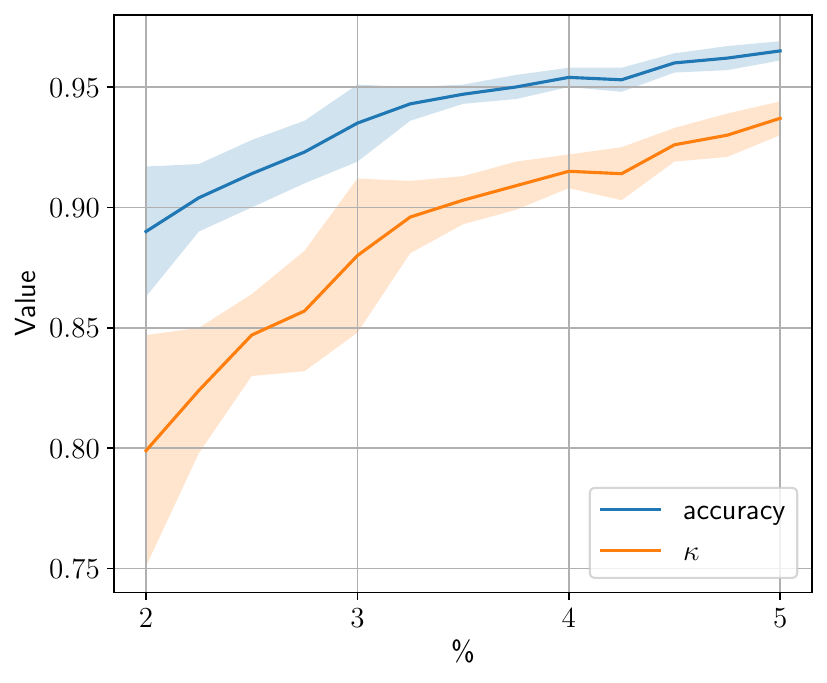}}
\subfloat[{\em Protozoan cysts}]{\includegraphics[width=0.33\textwidth]{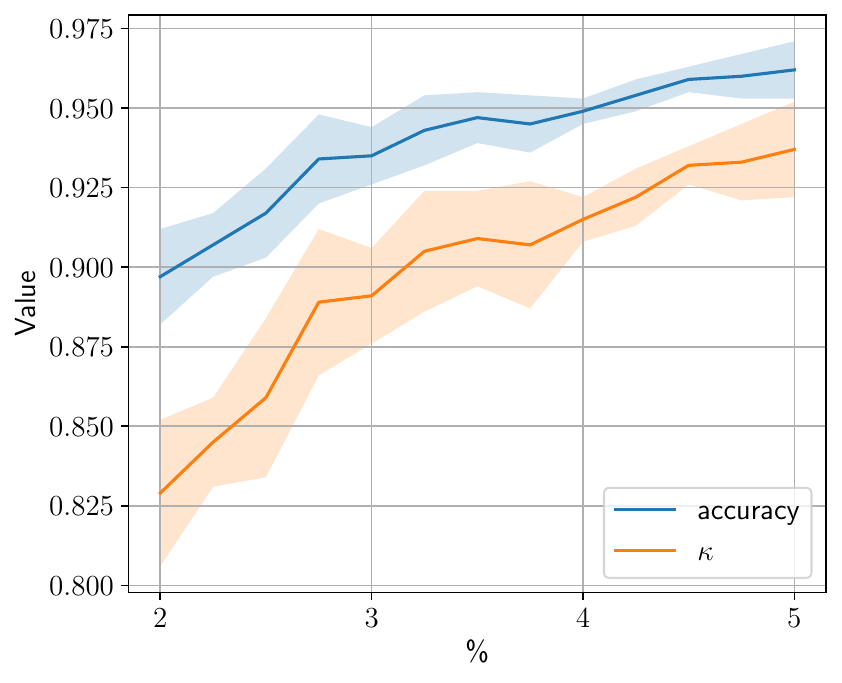}}
\caption{Plots of the test results showing the mean and standard deviation of accuracy and $\kappa$ for active-DeepFA across different percentage values of labeled data during active learning expansion.}
\label{fig:percentages}
\end{figure*}

Most semi-supervised methods rely on a fixed reduced set of labeled samples to train a network in conditions of labeled data shortage. Previous DeepFA-based approaches also follow this strategy~\cite{benato2023deep,aparco2024contrastive,aparco2025consensus}. This reduced set is often sampled randomly and then submitted to an expert for labeling. However, this sampling scheme overlooks the selection of informative samples that could enhance the learning process while lessening the annotation workload. The active-DeepFA method builds on this insight by systematically retrieving the samples whose pseudo-labels have the lowest labeling confidence value, derived after label propagation by OPFSemi. These samples exhibit the characteristic of being the most difficult to classify. Figure~\ref{fig:certainty} shows the heatmap of the labeling confidence values following OPFSemi's label propagation on the 2D projection of the {\em Helminth egg}'s output deep features. It can be seen that samples whose pseudo-labels have the lowest confidence values lie in decision boundary regions, where samples of different classes are either in proximity or overlapping. Therefore, it is reasonable to infer that these samples contain meaningful information, which can be leveraged to improve the generalization performance of the networks. This assumption is validated by the results achieving better classification scores than its counterparts with the same amount of labeled samples $|{\cal Z}_{L}|=5\%\cdot|{\cal Z}_{D}|$.

\begin{table}[!htb]
    \caption{Test results of accuracy and $\kappa$ for active-DeepFA over different percentages values of labeled data during active learning expansion.}
    \centering
    \renewcommand{\arraystretch}{1.2}
    \resizebox{.7\columnwidth}{!}{%
    \begin{tabular}{ c | c | c | c | c}
        \multirow{2}{*}{$|{\cal Z}_{L}|$} & \multirow{2}{*}{\bf Metric} & \multicolumn{3}{c}{\bf Datasets} \\
        \cline{3-5}
        & & {\it Helminth larvae} & {\it Helminth eggs} & {\it Protozoan cysts} \\
        \hline
        \multirow{2}{*}{$2\%\cdot|{\cal Z}_{D}|$} & accuracy & 0.953 $\pm$ 0.019 & 0.890 $\pm$ 0.027 & 0.897 $\pm$ 0.015 \\
        & $\kappa$ & 0.802 $\pm$ 0.076 & 0.799 $\pm$ 0.048 & 0.829 $\pm$ 0.023 \\
        \hline
        \multirow{2}{*}{$3\%\cdot|{\cal Z}_{D}|$} & accuracy & 0.975 $\pm$ 0.009 & 0.935 $\pm$ 0.016 & 0.935 $\pm$ 0.009 \\
        & $\kappa$ & 0.889 $\pm$ 0.042 & 0.880 $\pm$ 0.032 & 0.881 $\pm$ 0.015 \\
        \hline
        \multirow{2}{*}{$4\%\cdot|{\cal Z}_{D}|$} & accuracy & 0.983 $\pm$ 0.005 & 0.954 $\pm$ 0.004 & 0.949 $\pm$ 0.004 \\
        & $\kappa$ & 0.922 $\pm$ 0.026 & 0.915 $\pm$ 0.007 & 0.915 $\pm$ 0.007 \\
        \hline
        \multirow{2}{*}{$5\%\cdot|{\cal Z}_{D}|$} & accuracy & \textbff{0.992 $\pm$ 0.003} & \textbff{0.965 $\pm$ 0.004} & \textbff{0.962 $\pm$ 0.009} \\
        & $\kappa$ & \textbff{0.959 $\pm$ 0.013} & \textbff{0.937 $\pm$ 0.007} & \textbff{0.937 $\pm$ 0.015} \\
        \hline
    \end{tabular} }
    \label{tab:percentages}
\end{table}

Table~\ref{tab:percentages} and Figure~\ref{fig:percentages} display the test results for different percentages of labeled data after the active learning expansion phase. It is observed that comparable results to the baselines are attained with $|{\cal Z}_{L}|=3\%\cdot|{\cal Z}_{D}|$ for the {\em Helminth larvae} and {\em Protozoan cysts} datasets, and with $|{\cal Z}_{L}|=4\%\cdot|{\cal Z}_{D}|$ for the {\em Helminth eggs} dataset. Therefore, we show that the active learning procedure effectively reduces the annotation effort by requiring a smaller number of labeled samples to obtain similar results to those of the baselines. 

In addition, the recurrent expansion of the labeled set through active learning has a critical impact on various components of the active-DeepFA method. The inclusion of these meaningful samples incrementally enhance the low-dimensional space learned by the contrastive projection head $P_{c}$ via the minimization of ${\cal L}_{S}^{cl}$, which consequently improves the representation ability of the encoder $E$.  Moreover, these meaningful samples gradually boost the feature space learned by the classification projection head $P_{x}$, which minimizes both ${\cal L}_{S}$ and ${\cal L}_{U}^{ssl}$. Furthermore, during the label propagation phase, this set of samples compose the set of prototypes from which OPFSemi propagates the labels. This progressively improves the feature space of the encoder $E$ as epochs evolve, which is demonstrated by the results in Table~\ref{tab:percentages}. 

\begin{table*}[tb]
    \caption{Evolution of the 2D projection of the output deep features of encoder $E$ across incremental values of percentages of labeled data as the labeled set expands through active learning.}
    \label{tab:projection_evol}
    \centering
    \renewcommand{\arraystretch}{1.25}
    \resizebox{\columnwidth}{!}{%
    \begin{tabular}{
    >{\centering\arraybackslash} m{2.3cm} | >{\centering\arraybackslash} m{4.3cm} | >{\centering\arraybackslash} m{4.3cm} | >{\centering\arraybackslash} m{4.3cm} | >{\centering\arraybackslash} m{4.3cm}
    }
        \multirow{2}{*}{\bf Dataset} & \multicolumn{4}{c}{$|{\cal Z}_{L}|$} \\
        \cline{2-5}
        & $2\%\cdot|{\cal Z}_{D}|$ & $3\%\cdot|{\cal Z}_{D}|$ & $4\%\cdot|{\cal Z}_{D}|$ & $5\%\cdot|{\cal Z}_{D}|$ \\
        \hline
        {\em Helminth larvae} (2 classes)
        & \includegraphics[width=0.25\textwidth]{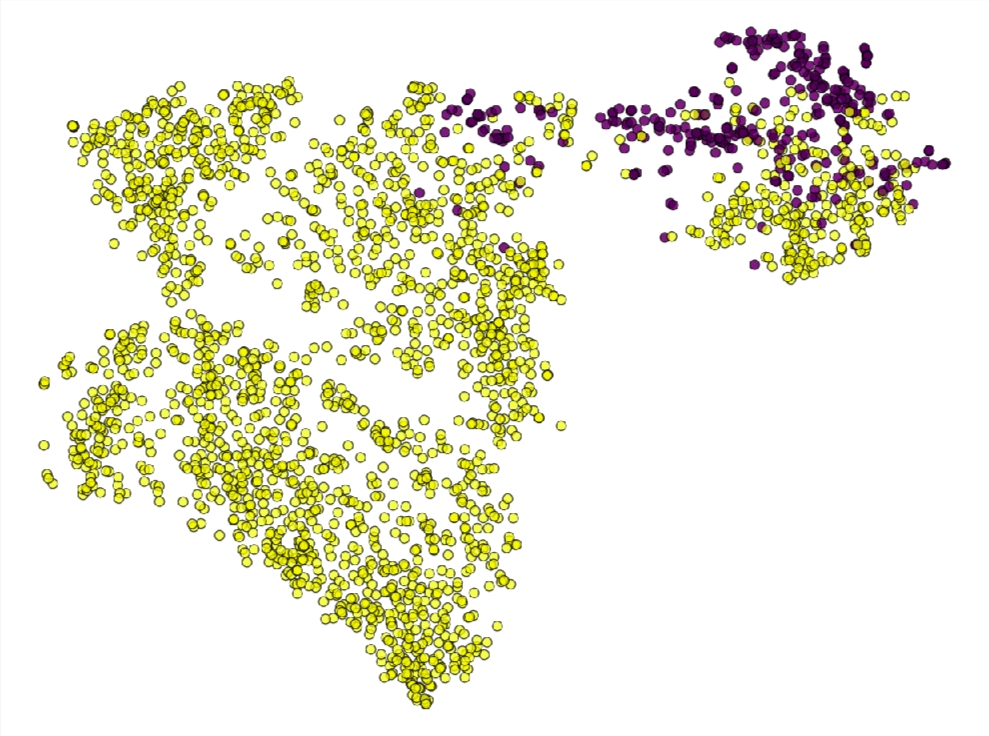} 
        & \includegraphics[width=0.25\textwidth]{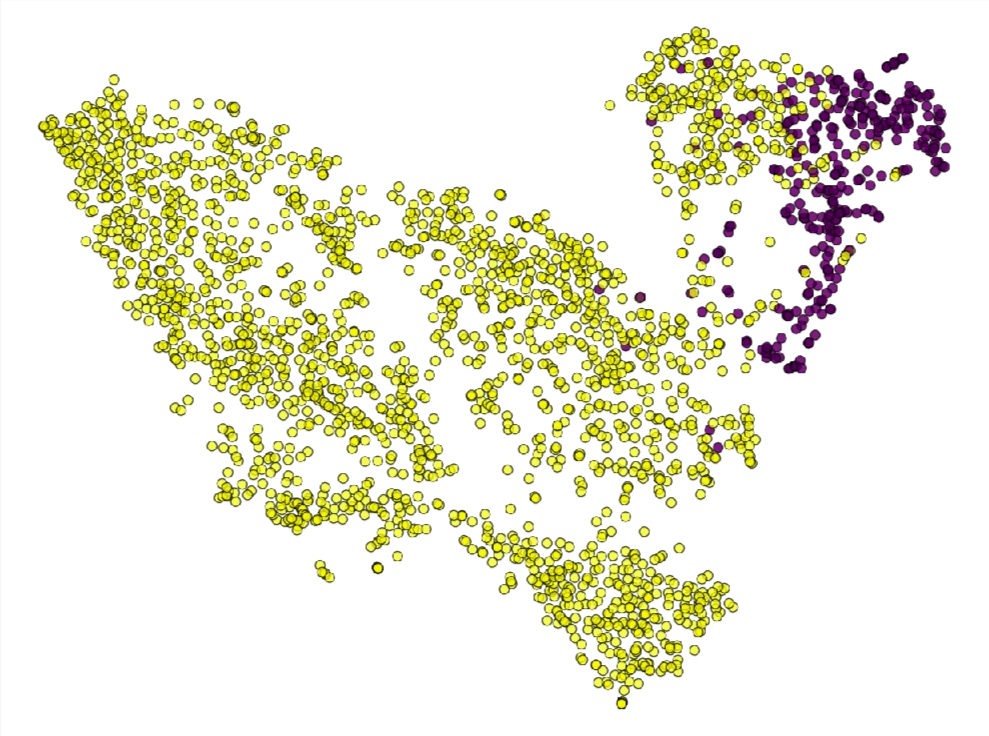} 
        & \includegraphics[width=0.25\textwidth]{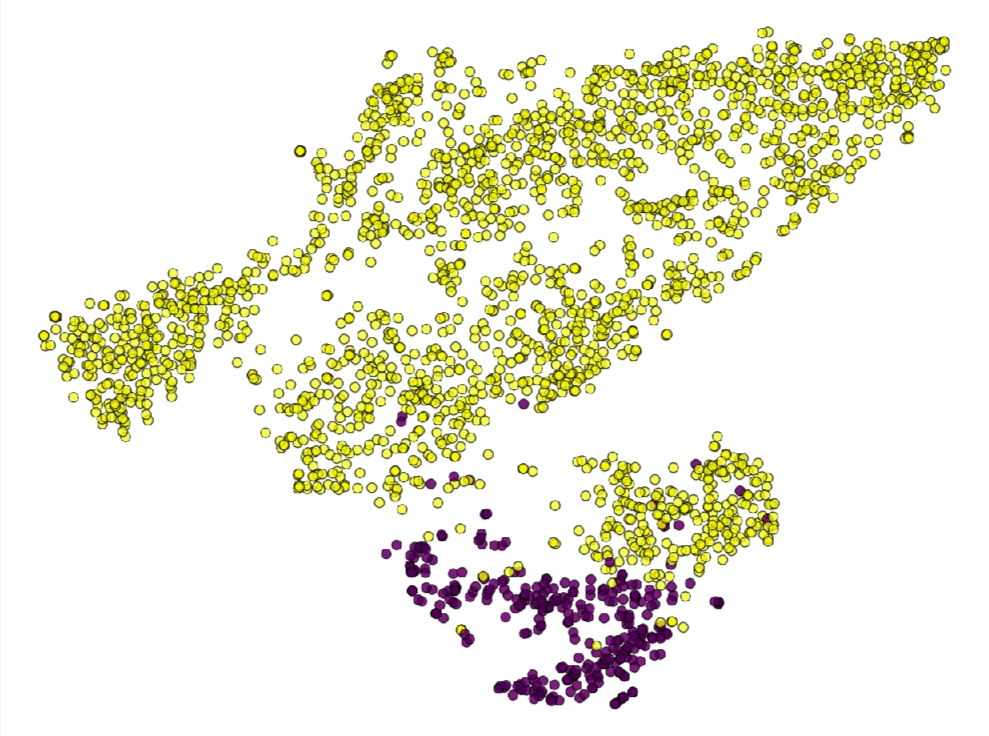}  
        & \includegraphics[width=0.25\textwidth]{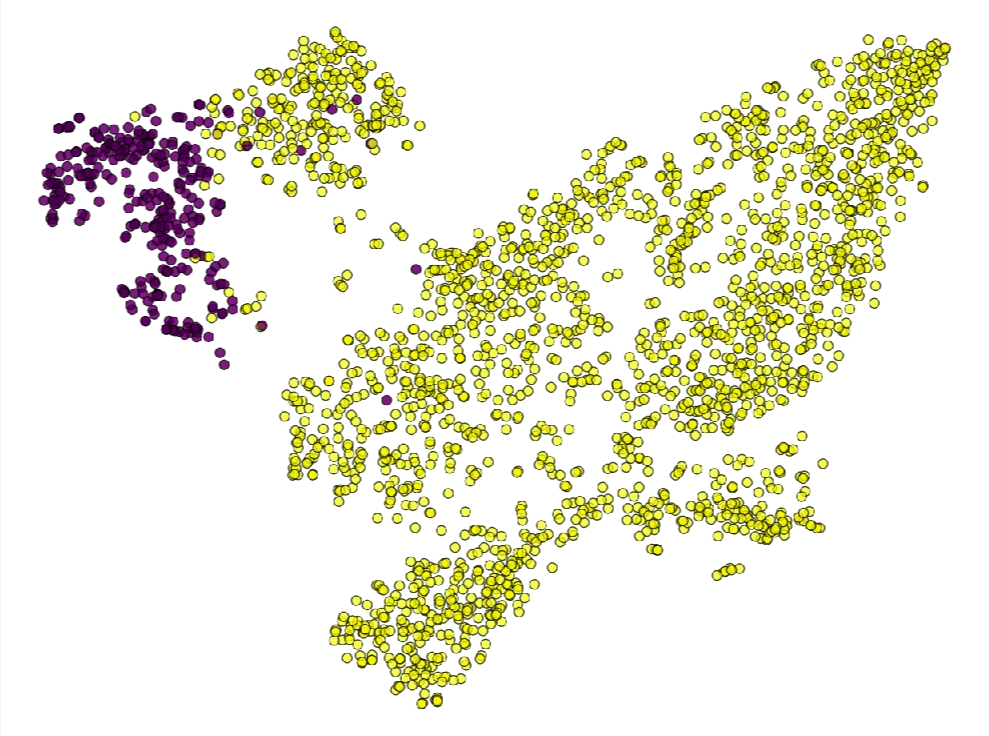} 
        \\
        \hline
        {\em Helminth eggs} (9 classes)
        & \includegraphics[width=0.25\textwidth]{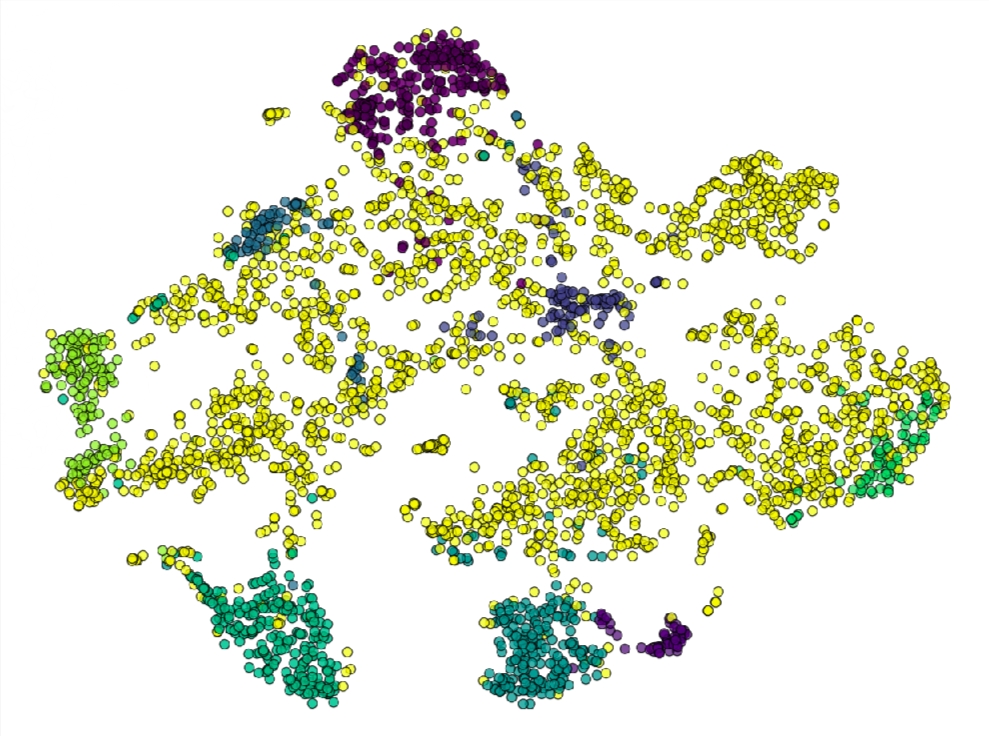} 
        & \includegraphics[width=0.25\textwidth]{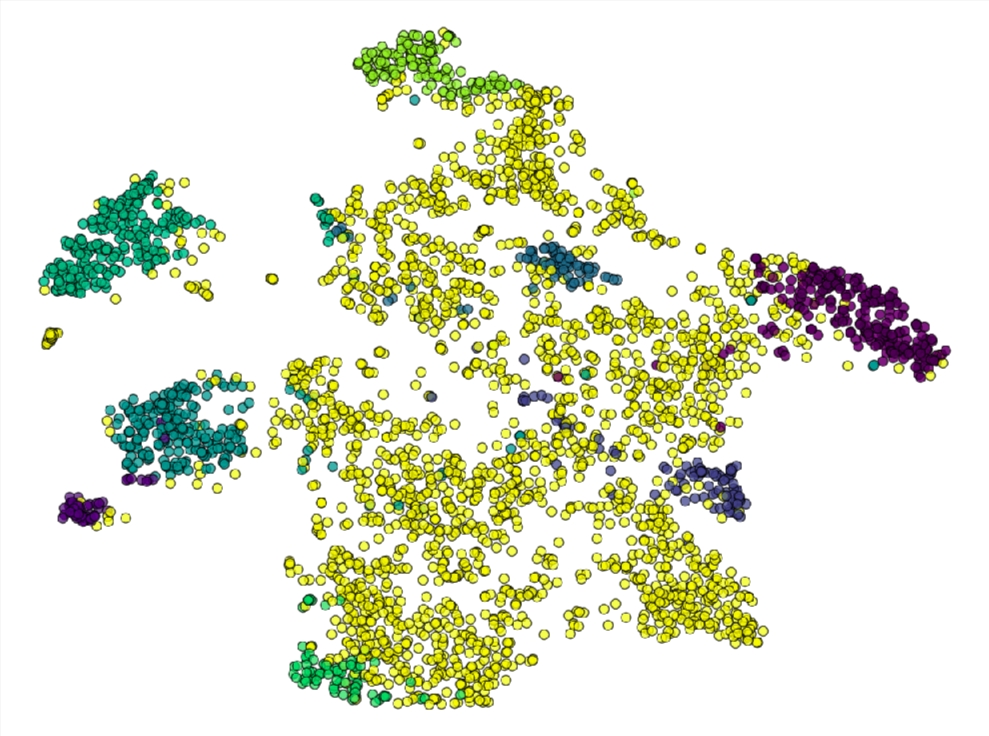} 
        & \includegraphics[width=0.25\textwidth]{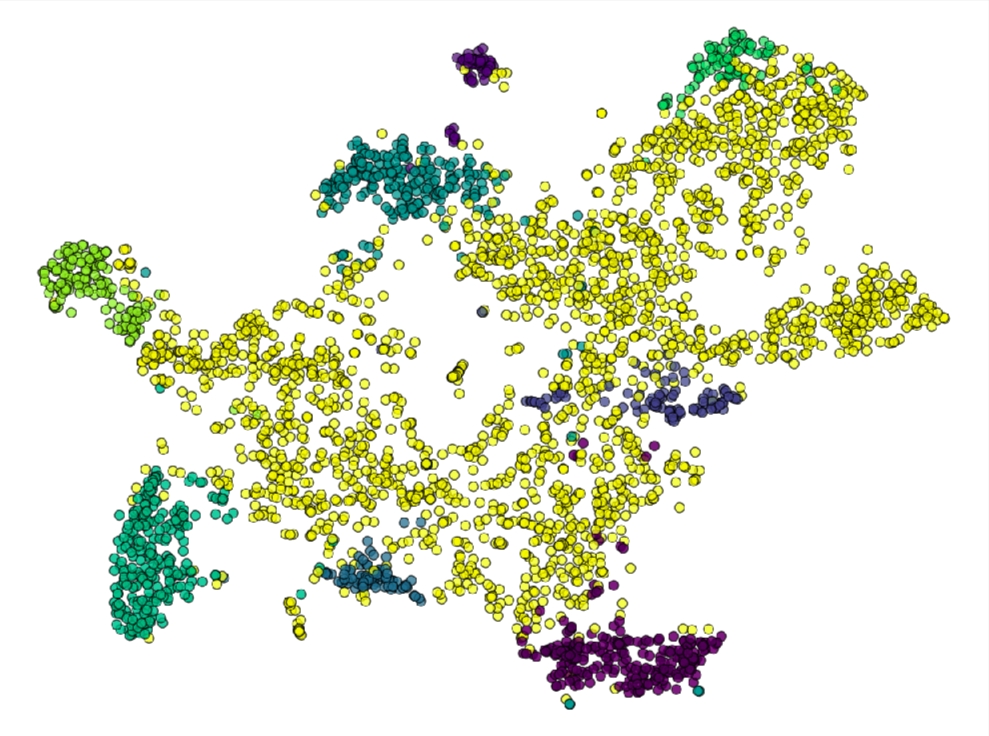}  
        & \includegraphics[width=0.25\textwidth]{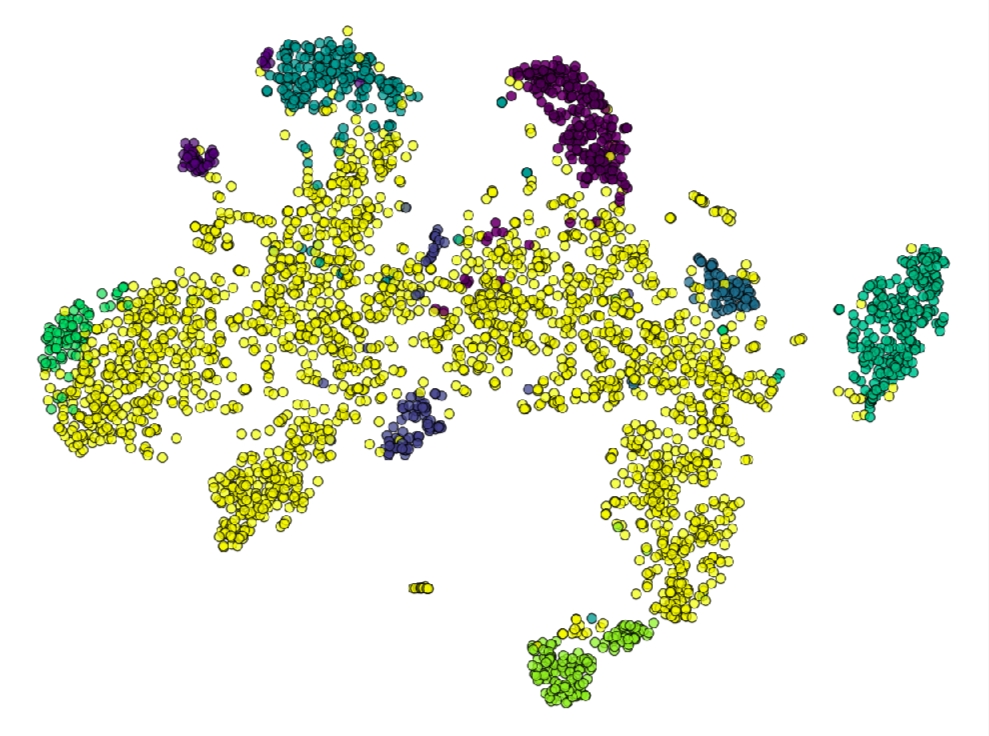}
        \\
        \hline
        {\em Protozoan cysts} (7 classes)
        & \includegraphics[width=0.25\textwidth]{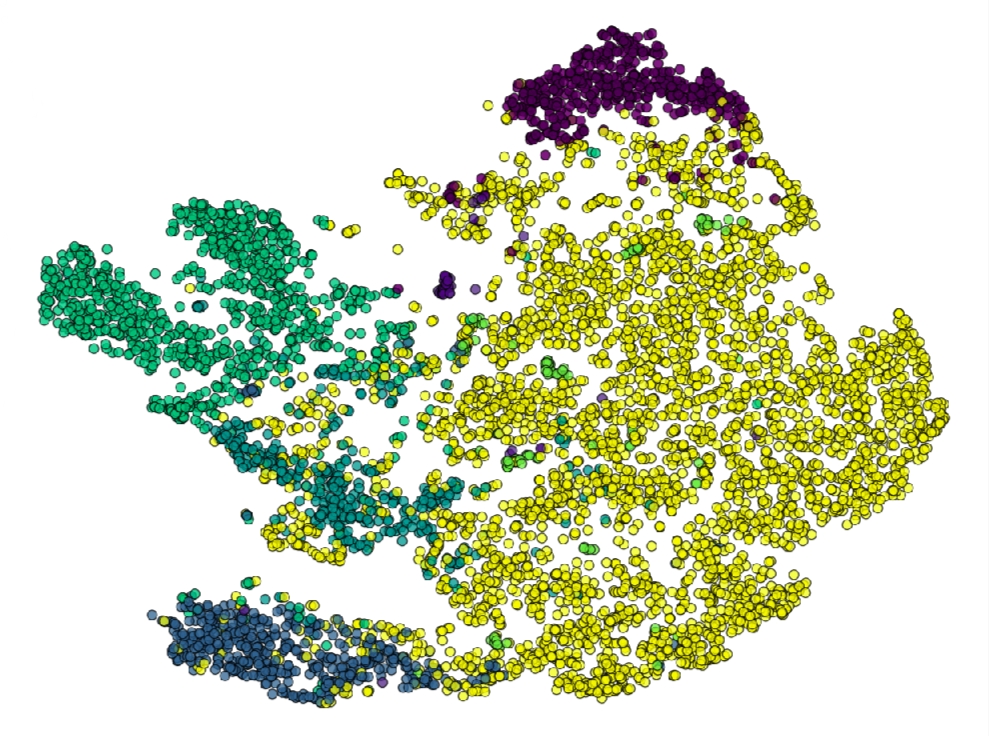} 
        & \includegraphics[width=0.25\textwidth]{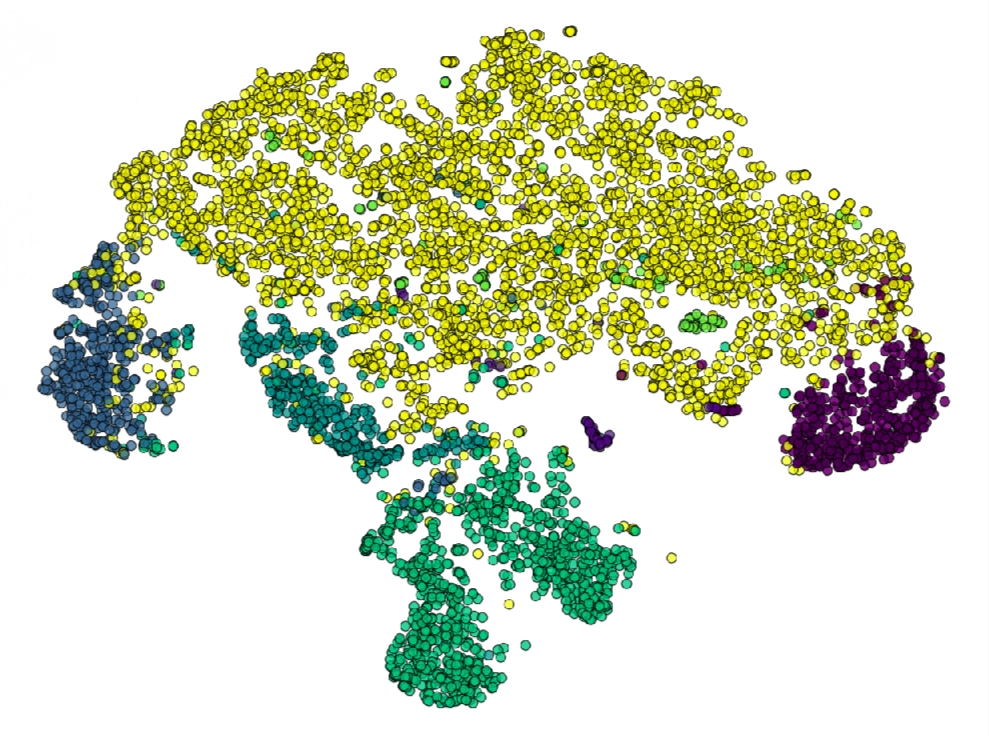} 
        & \includegraphics[width=0.25\textwidth]{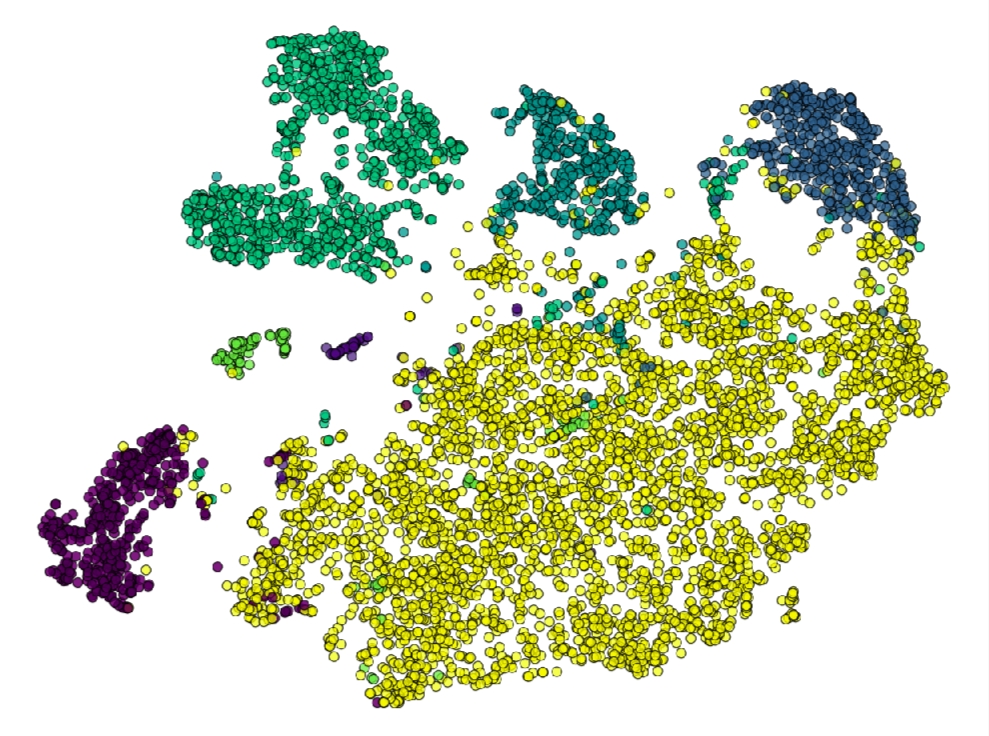}  
        & \includegraphics[width=0.25\textwidth]{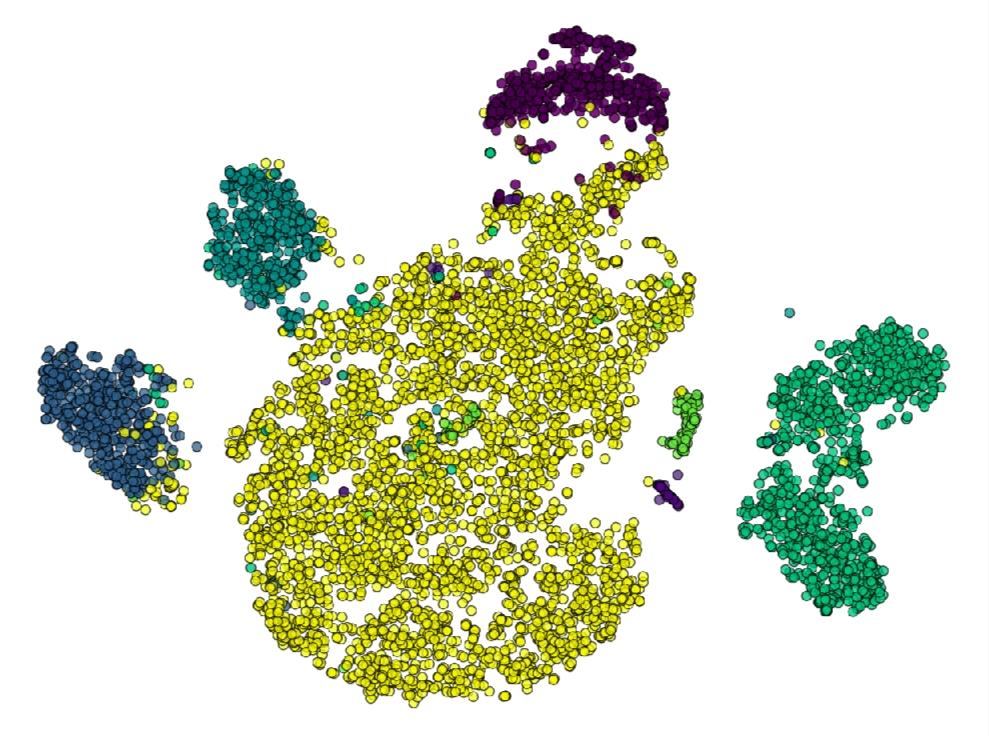}
    \end{tabular}}
\end{table*}

\subsubsection{Influence of cross-training}

The integration of two collaborative networks within a co-training scheme into the DeepFA framework plays a pivotal role in mitigating confirmation bias and overfitting. This is achieved by periodically exposing the networks to new information through the mutual exchange of reliable pseudo-labels. By doing this, the networks prevent overfitting the pseudo-labels generated from the 2D projection of the output feature space of its own encoder, which can be detrimental to the learning process if the pseudo-labels are inaccurate. This is revealed by the results in Table~\ref{tab:baselines}, where the methods that implement this strategy --Wang~\etal~\cite{wang2021sar}, cont-DeepFA and active-DeepFA -- achieve better generalization performance on the test set than conf-DeeFA, which trains the network on the pseudo-labels generated from the 2D projection of its own feature space.

During the active learning procedure, each network provides its own set of least-confident pseudo-labels. We capitalize on these two different views of the data by joining the sets and selecting the most uncertain pseudo-labels, thus retrieving more informative samples and enhancing the learning process. Also, active-DeepFA presents an improvement in terms of storage efficiency over its counterparts due to being run in a single iteration and thus producing only two networks. In contrast, Wang~\etal~\cite{wang2021sar} and cont-DeepFA generate a number of networks equal to twice the number of iterations.

\subsubsection{Influence of label propagation by OPFSemi}

The performance of OPFSemi is boosted by conducting label propagation on two distinct data representations derived from the output feature space of the networks' encoders. The periodic exchange of the most reliable pseudo-labeled samples between the networks improves the feature space as epochs proceed. The 2D projection of this feature space progressively exhibits more defined clusters comprising samples of the same class (see Table~\ref{tab:projection_evol}). This provides suitable projection conditions for OPFSemi, enhancing the label propagation procedure and yielding more accurate pseudo-labels. Table~\ref{tab:projection_evol} shows the evolution of the encoder's output feature space for increasing values of percentages of labeled data as the labeled set expands through active learning. It can be seen that as epochs advance, samples of the same class (color) form more separated and compact clusters, which improves the classification metrics, as evidenced by the results in Table~\ref{tab:percentages}. The aforementioned effect can also be attributed in part to the samples periodically labeled by active learning that act as prototypes. These samples are regarded as the most difficult to classify and their 2D projections are in close proximity to decision boundaries. Therefore, their inclusion in the set of prototypes during label propagation encourages the production of better-defined and more separated clusters. 

In addition, the integration of the OPFSemi classifier in the co-training setup, as implemented in both cont-DeepFA and active-DeepFA, shows an improvement over the generalization performance of~\cite{wang2021sar} (see Table~\ref{tab:baselines}), which conducts pseudo-labeling in a self-training way using the same networks to assign pseudo-labels to the unlabeled samples. This result reveals the effectiveness of OPFSemi as an effective tool for label propagation, providing reliable pseudo-labels in appropriate projection scenarios. Moreover, the application of strong augmentations on the images from the sets ${\cal Z}_{U}^{pl(*)}$, $*\in\{1,2\}$, during the training of the networks by minimizing ${\cal L}_{U}^{ssl}$, acts as a key factor in achieving better generalization ability. This is validated by the results shown in Table~\ref{tab:baselines}.

\subsection{Comparison with state-of-the-art DSSL methods}

This section focuses on benchmarking active-DeepFA with other state-of-the-art (SOTA) DSSL methods from the literature. We selected six SOTA DSSL methods categorized into three distinct groups based on their underlying solution strategy: {\emph Pseudo-labeling}: Pseudo-label~\cite{lee2013pseudo}, {\em Consistency Regularization}: $\Pi$-Model~\cite{laine2017temporal}, Mean Teacher~\cite{tarvainen2017mean}, VAT~\cite{miyato2018virtual} and UDA~\cite{xie2020unsupervised}, and {\em Hybrid}: FixMatch~\cite{sohn2020fixmatch}. We employ the Unified SSL Benchmark (USB) library~\cite{usb2022} that contains a Pytorch implementation of the aforesaid DSSL methods. All six DSSL methods use $|{\cal Z}_{L}|=5\%\cdot|{\cal Z}_{D}|$ from the outset. In addition, the labeled set ${\cal Z}_{L}$ is used as validation set since all methods rely on it to select the best-performing model. 

\begin{table}[htb!]
    \caption{Test results of accuracy and $\kappa$ for state-of-the-art DSSL methods with $|{\cal Z}_{L}|=5\%\cdot|{\cal Z}_{D}|$.}
    \centering
    \renewcommand{\arraystretch}{1.2}
    \resizebox{.7\columnwidth}{!}{%
    \begin{tabular}{ l | c | c | c | c}
        \multirow{2}{*}{\bf Method} & \multirow{2}{*}{\bf Metric} & \multicolumn{3}{c}{\bf Datasets} \\
        \cline{3-5}
        & & {\it Helminth larvae} & {\it Helminth eggs} & {\it Protozoan cysts} \\
        \hline
        \multirow{2}{*}{Pseudo-label~\cite{lee2013pseudo}} & accuracy & 0.960 $\pm$ 0.010 & 0.916 $\pm$ 0.008 & 0.903 $\pm$ 0.011 \\
        & $\kappa$ & 0.811 $\pm$ 0.057 & 0.850 $\pm$ 0.008 & 0.836 $\pm$ 0.023 \\
        \hline
        \multirow{2}{*}{$\Pi$-Model~\cite{laine2017temporal}} & accuracy & 0.962 $\pm$ 0.020 & 0.922 $\pm$ 0.003 & 0.912 $\pm$ 0.009 \\
        & $\kappa$ & 0.836 $\pm$ 0.081 & 0.855 $\pm$ 0.004 & 0.853 $\pm$ 0.016 \\
        \hline
        \multirow{2}{*}{Mean Teacher~\cite{tarvainen2017mean}} & accuracy & 0.969 $\pm$ 0.009 & 0.928 $\pm$ 0.017 & 0.906 $\pm$ 0.003 \\
        & $\kappa$ & 0.860 $\pm$ 0.037 & 0.870 $\pm$ 0.033 & 0.843 $\pm$ 0.005 \\
        \hline
        \multirow{2}{*}{VAT~\cite{miyato2018virtual}} & accuracy & 0.965 $\pm$ 0.006 & 0.933 $\pm$ 0.013 & 0.912 $\pm$ 0.010 \\
        & $\kappa$ & 0.844 $\pm$ 0.031 & 0.880 $\pm$ 0.021 & 0.854 $\pm$ 0.018 \\
        \hline
        \multirow{2}{*}{UDA~\cite{xie2020unsupervised}} & accuracy & 0.966 $\pm$ 0.009 & 0.933 $\pm$ 0.002 & 0.914 $\pm$ 0.004 \\
        & $\kappa$ & 0.851 $\pm$ 0.033 & 0.880 $\pm$ 0.003 & 0.857 $\pm$ 0.007 \\
        \hline
        \multirow{2}{*}{FixMatch~\cite{sohn2020fixmatch}} & accuracy & 0.968 $\pm$ 0.015 & 0.925 $\pm$ 0.011 & 0.914 $\pm$ 0.006 \\
        & $\kappa$ & 0.857 $\pm$ 0.060 & 0.862 $\pm$ 0.020 & 0.856 $\pm$ 0.011 \\
        \hline
        \multirow{2}{*}{active-DeepFA} & accuracy & \textbff{0.992 $\pm$ 0.003} & \textbff{0.965 $\pm$ 0.004} & \textbff{0.962 $\pm$ 0.009} \\
        & $\kappa$ & \textbff{0.959 $\pm$ 0.013} & \textbff{0.937 $\pm$ 0.007} & \textbff{0.937 $\pm$ 0.015} \\
        \hline
    \end{tabular} }
    \label{tab:dssl_methods}
\end{table}

Table~\ref{tab:dssl_methods} shows the mean and standard deviation of both accuracy and $\kappa$ for all six methods. It can be observed that active-DeepFA excels by consistently surpassing all its counterparts in both metrics across the three datasets. From Tables~\ref{tab:percentages} and~\ref{tab:dssl_methods}, it can be seen that with only $|{\cal Z}_{L}|=3\%\cdot|{\cal Z}_{D}|$, active-DeepFA -- as with the baselines -- achieves results comparable to those of the state-of-the-art methods, therefore further validating its efficacy in reducing the annotation burden while attaining satisfactory results. 

In particular, active-DeepFA demonstrates evident superiority over Pseudo-label, pointing that the meta-pseudo-labeling approach within a teacher-student setting enhances the conventional self-training pseudo-labeling scheme where the network itself assigns pseudo-labels to unlabeled data. Also, this result supports the role of the OPFSemi classifier as teacher within the teacher-student paradigm. On the other hand, consistency regularization-based methods improve the results achieved by Pseudo-label across all three datasets, suggesting that incorporating data augmentation techniques and consistency constraints into the objective loss function could further improve the effectiveness of our method. It can also be noted that Fixmatch, a hybrid approach that leverages on both self-training pseudo-labeling and consistency regularization, achieves consistent results across all datasets, thus suggesting that incorporating a teacher (\eg, OPFSemi) within a teacher-student setup could enhance its overall performance. The comparison between active-DeepFA and Mean Teacher, another teacher-student-based approach, highlights the superiority of teacher-student-based pseudo-labeling within a co-training setting. The active-DeepFA method invariably demonstrates superior performance over its consistency regularization counterparts, further substantiating its effectiveness and validating the efficacy of the proposed active learning strategy. An additional point to emphasize is that the unbalanced composition of the datasets coupled with the pronounced similarity among certain classes -- particularly between impurities and parasites -- can account for the divergence observed across the results produced by different methods. In this context, active-DeepFA proves its robustness and flexibility in addressing new data yielding consistent performance across all three datasets.

\section{Final Remarks}
\label{sec:final_remarks}

In the present work, we introduced active-DeepFA, an active contrastive-based meta-pseudo-labeling approach that follows a teacher-student paradigm within a co-training setup. It enables the training of non-pretrained custom CNN architectures for image classification in conditions of short supply of labeled and abundance of unlabeled data. It achieves this by incorporating deep feature annotation (DeepFA) into a cross-training procedure that implements two collaborative networks, which in turn minimize an objective loss function comprising three components: {\it i}) a weighted supervised contrastive loss, {\it ii}) a supervised loss, and {\it iii}) a semi-supervised loss. Label propagation via OPFSemi and labeled set expansion by active learning are performed at regular epoch intervals. Then, the most reliable pseudo-labels are submitted to the networks for retraining in a cross-training manner, mitigating confirmation bias and overfitting while improving the networks' generalization performance as epochs proceed. The active-DeepFA method capitalizes on supervised contrastive learning to enhance the networks' representation ability and to circumvent the prerequisite of pre-trained CNN architectures. It leverages active learning by periodically retrieving the most uncertain/informative samples identified through label propagation and submitting them to an oracle for annotation. This labeled set expansion has a critical impact on the three components of the objective loss function. The expanded labeled set improves the training set for both supervised contrastive and supervised learning. Also, it enhances the set of prototypes for label propagation via OPFSemi generating more reliable pseudo-labels and thus boosting semi-supervised learning. We assessed active-DeepFA on three challenging unbalanced biological image datasets. The labeled data scarcity scenario is emulated by randomly labeling only 1\% of the samples of each dataset in a stratified fashion. The labeled set is incrementally expanded at regular epoch intervals through active learning until 5\% of the dataset is labeled. We also evaluated active-DeepFA with intermediate percentage values of expanded labeled data during the active learning procedure. The active-DeepFA method is compared to three direct baselines and to six other state-of-the-art DSSL methods that fall into three different categories according to their solution approach. Our method notably improves its baselines and outperforms the state-of-the-art methods for all three datasets with 5\% of labeled data, thus demonstrating its effectiveness and robustness. Moreover, it achieves comparable results to those of its counterparts with only 3\% of labeled data, proving their efficacy in reducing the annotation effort while achieving satisfactory results. 

As future work, we intend to improve the results here presented by incorporating data augmentation strategies and consistency constraints into the objective loss function to enhance the generalization performance of the models and further reduce the quantity of labeled data to yield satisfactory results. Moreover, we plan to replace the 2D projection of t-SNE by using the low-dimensional output of a projection head to make the method faster.

\section*{Declarations}

\subsection*{Contributions}
DA, AXF, and PJR contributed to the conception of this study. DA performed the experiments. JFG contributed to the data curation. AXF and PJR contributed to the supervision. DA wrote the original draft. DA, AXF, and PJR contributed to the review and editing. DA is the main contributor and writer of this manuscript. All authors read and approved the final manuscript.


\subsection*{Competing Interests}
The authors declare that they have no competing interests.


\subsection*{Funding}
 The authors acknowledge grants from {\it Santander Bank Brazil}; 
 CAPES/STIC-AMSUD,%
 ~\#88887.878869/2023-00;
 CNPq,%
 ~\#140930/2021-3, 
 ~\#314293/2023-0, 
 ~\#313329/2020-6, 
 ~\#442950/2023-3, 
 ~\#304711/2023-3; 
 FAPESP, %
 \#2023/14427-8, 
 \#2023/04318-7, 
 \#2014/12236-1, 
 \#2013/07375-0. 

\subsection*{Availability of data and materials}
The datasets analyzed during the current study are available at \url{https://github.com/LIDS-UNICAMP/intestinal-parasites-datasets}.

\bibliographystyle{unsrt}  
\bibliography{references}  






\end{document}